\def\year{2021}\relax
\documentclass[letterpaper]{article} 
\usepackage{aaai21}  
\usepackage{times}  
\usepackage{helvet} 
\usepackage{courier}  
\usepackage[hyphens]{url}  
\usepackage{graphicx} 
\def\UrlFont{\rm}  
\usepackage{natbib}  
\usepackage{caption} 
\renewcommand{\UrlFont}{\ttfamily\small}

\usepackage{dirtytalk}
\usepackage{booktabs}

\usepackage{graphicx}
\usepackage{lipsum, color}
\usepackage{multirow}
\usepackage{array}
\usepackage{multicol}
\usepackage{url}
\makeatletter
\g@addto@macro{\UrlBreaks}{\UrlOrds}
\makeatother
\usepackage{array}
\usepackage{graphbox}
\newcolumntype{L}[1]{>{\raggedright\let\newline\\\arraybackslash\hspace{0pt}}m{#1}}
\newcolumntype{C}[1]{>{\centering\let\newline\\\arraybackslash\hspace{0pt}}m{#1}}
\newcolumntype{R}[1]{>{\raggedleft\let\newline\\\arraybackslash\hspace{0pt}}m{#1}}
\usepackage{pifont}
\usepackage{amsmath,amsfonts,amssymb,amsthm,bm}
\usepackage{mathtools}

\usepackage{caption}
\usepackage[inline]{enumitem}
\usepackage{caption, subcaption}
\usepackage{microtype}

\usepackage{xspace}
\usepackage{pgfplots}

\usepackage{microtype}
\usepackage{booktabs}

\usepackage{xspace}
\usepackage{xcolor}

\usepackage[colorinlistoftodos,prependcaption,textsize=scriptsize]{todonotes}

\newcommand{\textrank}{\textsc{Text\-Rank}\xspace}
\newcommand{\scenesum}{\textsc{Scene\-Sum}\xspace}
\newcommand{\graphtp}{\textsc{Graph\-T\-P}\xspace}

\definecolor{tp1}{HTML}{28B463}
\definecolor{tp2}{HTML}{196F3D}
\definecolor{tp3}{HTML}{808000}
\definecolor{tp4}{HTML}{FF0000}
\definecolor{tp5}{HTML}{CD6155}

 \newcommand{\attribute}[1]{\textsf{\fontsize{7.4pt}{.1pt}\selectfont{#1}}}

\title{Movie Summarization via Sparse Graph Construction}

\author{
    Pinelopi Papalampidi \quad
    Frank Keller \quad
    Mirella Lapata
    \\
}
\affiliations{
    Institute for Language, Cognition and Computation \\
    School of Informatics, University of Edinburgh \\

  {\tt p.papalampidi@sms.ed.ac.uk, \{keller,mlap\}@inf.ed.ac.uk }
}

\date{}

\begin{document}

\maketitle

\renewcommand{\UrlFont}{\ttfamily\scriptsize}

\begin{abstract}

  We summarize full-length movies by creating shorter videos
  containing their most informative scenes. We explore the hypothesis
  that a summary can be created by assembling scenes which are
  \emph{turning points} (TPs), i.e.,~key events in a movie that
  describe its storyline. We propose a model that identifies TP scenes
  by building a sparse movie graph that represents relations between
  scenes and is constructed using multimodal information\footnote{We
  make our data and code publicly available at
  \url{https://github.com/ppapalampidi/GraphTP}.}. According to
  human judges, the summaries created by our approach are more
  informative and complete, and receive higher ratings, than the
  outputs of sequence-based models and general-purpose summarization
  algorithms. The induced graphs are interpretable, displaying
  different topology for different movie genres.
  
\end{abstract}

\section{Introduction}
\label{sec:introduction}

Automatic summarization has received considerable attention due to its
importance for downstream applications. Although current research has
primarily focused on news articles
\cite{newsroom-naacl18,narayan-etal-2018-dont,liu-lapata-2019-hierarchical},
other application domains include meetings
\cite{murray2007automatic}, lectures
\cite{Fujii2007AutomaticEO}, social media \cite{syed2018task},
scientific articles \cite{teufel-moens-2002-articles}, and narratives
ranging from short stories
\cite{goyal-etal-2010-automatically,Finlayson:2012}
to books
\cite{mihalcea-ceylan-2007-explorations}, and movies
\cite{gorinski-lapata-2015-movie}.

\begin{figure}[t]
\centering
  \begin{small}
    \begin{tabular}{@{}p{8cm}@{}} \hline
                          \multicolumn{1}{c}{1. Opportunity} \\
      Introductory event that occurs after presentation of setting and
      background of main characters. (\attribute{Juno discovers she is pregnant
      with a child fathered by her friend and longtime
                          admirer.}) \\ \hline
      \multicolumn{1}{c}{2. Change of Plans} \\
     Main goal of story is defined; action begins
      to increase.
      (\attribute{Juno  decides to give the baby up for adoption.}) \\\hline
      \multicolumn{1}{c}{3. Point of No Return} \\
      Event that pushes the main characters to
      fully commit to their goal.
      (\attribute{Juno meets a couple, and agrees to a closed adoption.}) \\ \hline
      \multicolumn{1}{c}{4. Major Setback} \\ Event where everything falls apart,
      temporarily or
      permanently. (\attribute{Juno watches the couple's marriage fall apart.}) \\ \hline
      \multicolumn{1}{c}{5. Climax} \\ Final event of the main story, moment of resolution
      and ``biggest spoiler''. (\attribute{Juno gives birth and spouse from
      ex-couple claims newborn as single adoptive mother.}) \\\hline
\end{tabular}
\end{small}
\caption{\label{fig:TPs} Turning points (from the movie ``Juno'') and
  their definitions.}
\end{figure}

In this work, we aim at summarizing full-length movies by creating
shorter video summaries encapsulating their most informative
parts. Aside from enabling users to skim through movies quickly ---
Netflix alone has over 148 million subscribers worldwide, with more
than 6000--7000 movies, series, and shows available --- movie
summarization is an ideal platform for real-world natural language
understanding and the complex inferences associated with it. Movies
are often based on elaborate stories, with non-linear structure and
multiple characters, rendering the application of popular
summarization approaches based on position biases, importance, and
diversity problematic \cite{jung2019earlier}.  Another key challenge
in movie summarization lies in the scarcity of labeled data. For most
movies there are no naturally occurring summaries (trailers aim to
attract an audience to a film without revealing spoilers which a
summary will contain), and manually creating these would be a major
undertaking requiring substantial effort to collect, watch,
preprocess, and annotate videos. As a result, the majority of
available movie datasets contain at most a few hundred movies focusing
on tasks like Question-Answering (QA) or the alignment between video
clips and captions
\cite{tapaswi2016movieqa,xiong2019graph,rohrbach2015dataset}
which are limited to video snippets rather than entire movies, or
restricted to screenplays disregarding the video
\cite{gorinski-lapata-2015-movie,papalampidi2020screenplay}.

Following previous work \cite{gorinski-lapata-2015-movie}, we
formalize movie summarization as the selection of a few important
scenes from a movie. We further assume that important scenes display
events which determine the progression of the movie's narrative and
segment it into thematic sections. Screenwriting theory
\cite{thompson1999storytelling,cutting2016narrative,Hauge:2017} reserves the term \emph{turning points} (TPs) for
events which have specific functionality inside a narrative and reveal
its storyline. TPs are considered key for making successful movies
whose stories are expected to consist of six basic stages, defined by
five key turning points in the plot. An example of TPs and their
definitions is given in Figure~\ref{fig:TPs}.  Interestingly, TPs are
assumed to be the same, no matter the movie genre, and occupy the same
positions in the story (e.g., the Opportunity occurs after the first
10\% of a 90-minute comedy or a three-hour epic).

We propose that automatic movie summarization can be reduced to
turning point identification building on earlier work
\cite{Lehnert:1981,lohnert1981summarizing,mihalcea-ceylan-2007-explorations}
which claims that high level analysis is necessary for revealing
concepts central to a story. Although there exist several theories of
narrative structure \cite{cutting2016narrative}, we argue that turning
points are ideally suited to summarizing movies for at least three
reasons. Firstly, they are intuitive, and can be identified by naive
viewers \cite{papalampidi2019movie}, so there is hope the process can
be automated. Secondly, TPs have specific definitions and expected
positions which facilitate automatic identification especially in low
resource settings by providing prior knowledge (semantic and
positional). Thirdly, they provide data efficiency, since the
summarization problem is re-formulated as a scene-level classification
task and no additional resources are required for creating the movie
summaries over and above those developed for identifying turning
points.

We model TP identification (and by extension summarization) as a
supervised classification task.  However, we depart from previous
approaches to movie analysis which mostly focus on interactions
between \emph{characters}
\cite{do2018movie,tran2017exploiting,gorinski-lapata-2015-movie} and
model connections between \emph{events}. Moreover, we discard the
simplifying assumption that a screenplay consists of a \emph{sequence
  of scenes}
\cite{gorinski-lapata-2015-movie,papalampidi2019movie,papalampidi2020screenplay}
and instead represent interactions between scenes as a \emph{sparse
  graph}.  Specifically, we view the screenplay of a movie as a graph
whose nodes correspond to scenes (self-contained events) and edges
denote relations between them which we compute based on their
linguistic and audiovisual similarity.  In contrast to previous work
on general-purpose summarization that relies on fully connected graphs
\cite{mihalcea2004textrank,zheng2019sentence,wang-etal-2020-heterogeneous},
we induce sparse graphs by selecting a subset of nodes as neighbors
for a scene; the size of this subset is not set in advance but learnt
as part of the network.  Sparse graphs provide better
contextualization for scenes and tend to be more informative, as
different genres present different degrees of connectivity between
important events.  We rely on Graph Convolutional Networks (GCNs;
\citealt{duvenaud2015convolutional,kearnes2016molecular,kipf2017semi})
to encode relevant neighborhood information in the sparsified graph
for every scene which in turn contributes to deciding whether it acts
as a TP and should be included in the summary.

Our contributions can be summarized as follows: (a)~we approach movie
summarization directly via TP identification which we argue is a
well-defined and possibly less subjective task; (b)~we propose a TP
identification model which relies on sparse graphs and is constructed
based on multimodal information; (c) we find that the induced graphs
are meaningful with differing graph topologies corresponding to
different movie genres.

\section{Related Work}
\label{sec:related-work}

The computational treatment of narratives has assumed various guises
in the literature \cite{Mani:2012,Richards:ea:2009}. Previous work has
attempted to analyze stories by examining the sequence of events in
them \cite{Schank:Abelson:1975,chambers-jurafsky-2009-unsupervised},
plot units
\cite{mcintyre-lapata-2010-plot,goyal-etal-2010-automatically}
and their structure \cite{Lehnert:1981,Rumelhart:1980}, or the
interactions of characters in the narrative
\cite{black1979evaluation,Propp:1968,Vargas:ea:2014,Srivastava:ea:2016}.

The summarization of narratives has received less attention, possibly
due to the lack of annotated data for modeling and
evaluation. Nevertheless,
\citet{kazantseva-szpakowicz-2010-summarizing} summarize short stories
as a browsing aids to help users decide whether a story is interesting
to read. Other work
\cite{mihalcea-ceylan-2007-explorations,gorinski-lapata-2015-movie,tsoneva2007automated}
focuses on long-form narratives such as books or movies and adopts
primarily unsupervised, graph-based methods. More recently,
\citet{papalampidi2019movie} released TRIPOD, a dataset containing
screenplays and TP annotations and showed that TPs can be
automatically identified in movie narratives. In follow-on work,
\citet{papalampidi2020screenplay} further demonstrate that TPs provide
useful information when summarizing episodes from the TV series CSI.
In this work, we consider scenes as the basic summarization units, and
reduce the scene selection task to a TP identification problem.

Work on video understanding has also looked at movies. Existing
datasets \cite{tapaswi2016movieqa,rohrbach2015dataset} do not contain
more than a few hundred movies and focus mostly on \emph{isolated}
video clips rather than \emph{entire} narratives. For example,
\citet{tapaswi2015book2movie} align movie scenes to book chapters,
while \citet{xiong2019graph} align movie segments to descriptions
using a graph-based approach. \citet{rohrbach2015dataset} introduce a
dataset where video clips from movies are aligned to text descriptions
in order to address video captioning.  \citet{tapaswi2015aligning}
introduce a Question-Answering (QA) dataset based on movies, although
the questions are again restricted to isolated video
clips. \citet{frermann-etal-2018-whodunnit} analyze CSI episodes with
the aim of modeling how viewers identify the perpetrator. 

Our work is closest to \citet{papalampidi2019movie} in that we
also develop a model for identifying turning points in movies. While
they focus solely on textual analysis, we consider additional
modalities such as audio and video. Moreover, we model screenplays
more globally by representing them as graphs and inferring
relationships between scenes. Our graphs are interpretable and
differentially represent the morphology of different genres. Beyond
improving TP prediction, we further argue that narrative structure can
be \emph{directly} used to create video summaries for movies of any
genre. Previous work \cite{papalampidi2020screenplay} treats TPs as
latent representations with the aim of enhancing a supervised
summarization task. We do not assume goldstandard video summaries are
available, we claim that scenes which contain TPs can yield good enough
proxy summaries.

\begin{figure*}[t]
    \centering
    \includegraphics[width=\textwidth]{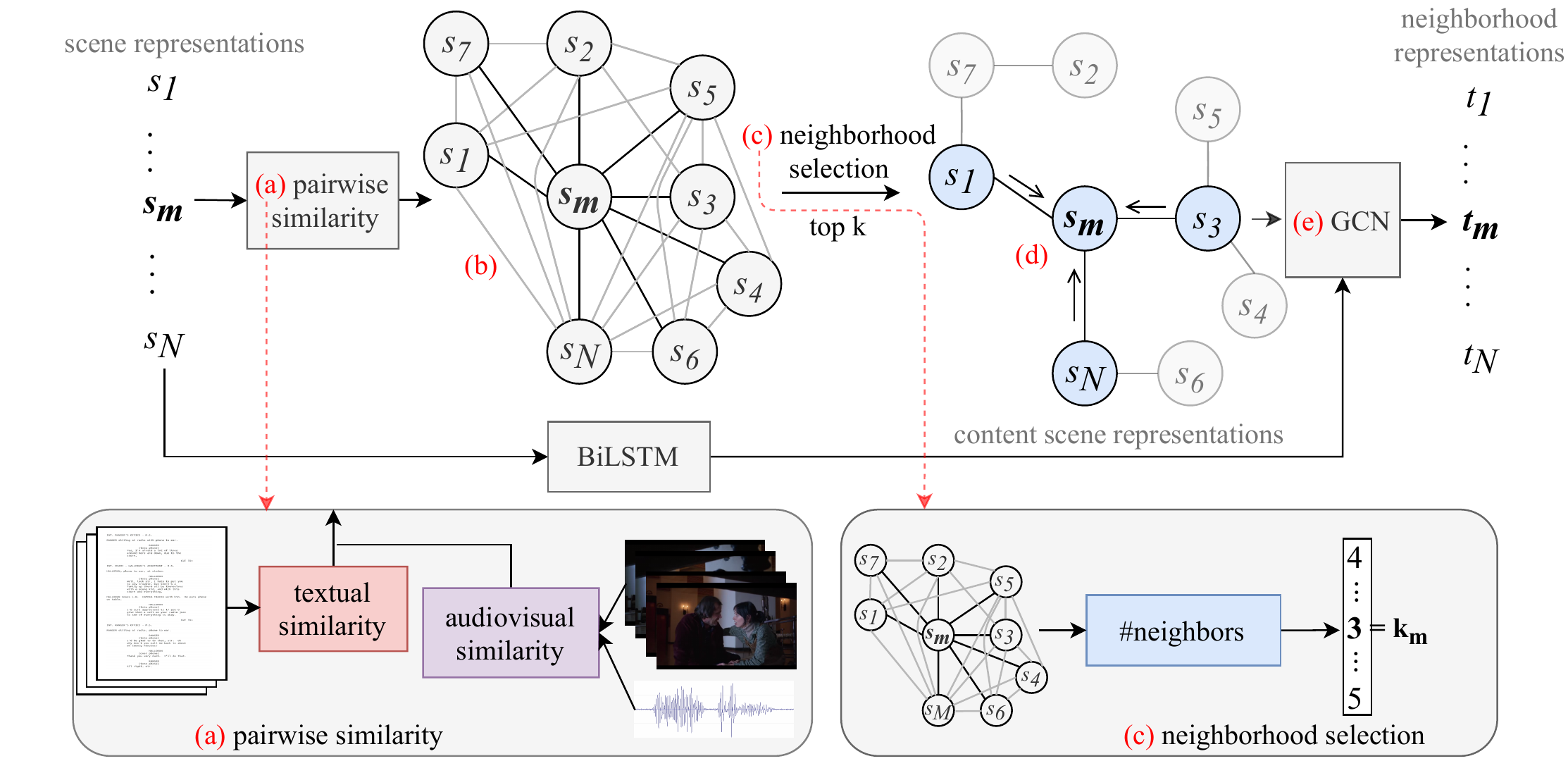}
    \caption{We construct a fully-connected graph based on pairwise
      textual and audiovisual similarity between scenes. The graph is
      sparsified (by automatically selecting the $k$ nearest neighbors
      per scene) and together with contextualized scene
      representations is fed to a one-layer GCN.}
    \label{fig:graph_tp}
\end{figure*}

\section{Problem Formulation}
\label{sec:problem-formulation}

Let~$\mathcal{D}$ denote a screenplay consisting of a sequence of
scenes $\mathcal{D}=\{s_1, s_2, \dots, s_n\}$. We aim at selecting a
smaller subset $\mathcal{D}' = \{s_i,\dots,s_k\}$ consisting of the
most \emph{informative} scenes describing the movie's
storyline. Hence, our objective is to assign a binary label~$y_i$ to
each scene~$s_i$ denoting whether it is part of the summary.

Furthermore, we hypothesize that we can construct an informative
summary by identifying TPs directly. As we explained earlier,
screenwriting theory \cite{Hauge:2017} postulates that most movie
narratives are delineated by five key events called turning points
(see Figure~\ref{fig:TPs}). Hence, we re-formulate the summarization
problem as follows: for each scene $s_{i} \in \mathcal{D}$ we assign a
binary label~$y_{it}$ denoting whether it represents turning
point~$t$. Specifically, we calculate probabilities
$p(y_{it} |s_i, \mathcal{D}, \theta)$ quantifying the extent to
which~$s_i$ acts as the $t^{th}$ TP, where $t \in [1,5]$ (and~$\theta$
are model parameters). During inference, we compose a summary by selecting
$l$~consecutive scenes that lie on the peak of the posterior
distribution
$\operatorname{argmax}\limits_{i=1}^{N}p(y_{it} |s_i, \mathcal{D},
\theta)$ for each TP.

\section{A Turning Point Graph Model}
\label{sec:turning-point-graph}

\subsection{Graph Construction}
\label{sec:graph-construction-2}

Let~$\mathcal{G}= (\mathcal{V}, \mathcal{E})$ denote a directed
screenplay graph with nodes $\mathcal{V}$ and
edges~$\mathcal{E}$. $\mathcal{G}$ consists of $N$~nodes, each
corresponding to a scene ($N$ varies with screenplay size; some
screenplays have many short scenes, while others only a few long
ones). We further represent~$\mathcal{G}$ by an adjacency matrix
$\mathcal{A} \in \mathcal{R}^{N \times N}$ where entry $a_{ij}$
denotes the weight of the edge from node~$i$ to node~$j$. We initially
construct a dense complete graph $\mathcal{G}$ with edge weights
representing the probability $p_{ij}$ of scene~$i$ being a neighbor of
scene~$j$ (see Figure~\ref{fig:graph_tp}(b)). We estimate $p_{ij}$ as:
\begin{equation}
        p_{ij} = \dfrac{\exp(e_{ij}/\tau)}{\sum_{t=1}^{T}
          \exp(e_{tj}/\tau)} 
          \label{eq:gumbel_softmax} 
\end{equation}
where $e_{ij}$ denotes the similarity between
scenes~$s_i$ and~$s_j$ (explained in the next section), $T$ are TPs
(see Figure~\ref{fig:TPs}), and~$\tau$ is a temperature parameter.

\paragraph{Similarity Computation} There are various ways to compute
the similarity~$e_{ij}$ between two scenes. In addition to linguistic
information based on the text of the screenplay, we wish to take
advantage of other modalities, such as audio and video.  Audiovisual
cues might be relatively superficial; simply on account of two scenes
sounding or seeming alike, it might not be possible to induce which
events are being described and their relations.  Nevertheless, we
expect audiovisual information to contribute to the similarity
computation by helping distinguish scenes which refer to the same
sub-story or event, e.g.,~because they have the same background, the
same characters, or similar noises. We thus express~$e_{ij}$ as a
composite term based mostly on textual information but also modulated by
audiovisual cues:
\begingroup\makeatletter\def\f@size{9.5}\check@mathfonts
\def\maketag@@@#1{\hbox{\m@th\large\normalfont#1}}%
\begin{gather} \label{eq:similarity}
 \hspace*{-.3cm}e_{ij} = u_{ij}  \bigg(\hspace*{-.1cm}\tanh(W_{i}v_{i} \hspace*{-.05cm}+\hspace*{-.05cm} b_{i})^\intercal
            \tanh(W_{j}v_{j} \hspace*{-.05cm}+ \hspace*{-.05cm}b_{j})\bigg)  \hspace*{-.05cm}+\hspace*{-.05cm} b_{ij}
\end{gather}\endgroup
where~$W_i$ and $W_j$ are weight matrices, $v_i$ and $v_j$ are
\emph{textual} vectors representing the content of scenes~$s_i$
and~$s_j$, and~$u_{ij}$ expresses the \emph{audiovisual}
similarity between~$s_i$ and $s_j$. If $u_{ij}$ is high,
the similarity between two scenes will be accentuated, but if it is
low when textual similarity is high, its influence will be modest (see
Figure~\ref{fig:graph_tp}(a)).

It is relatively straightforward to obtain textual representations for
scenes. The latter contain mostly dialogue (lines the actors speak) as
well as descriptions explaining what the camera sees. We first
calculate representations for the sentences included in a scene via a
pre-trained transformer-based sentence encoder
\cite{cer-etal-2018-universal}.  We then obtain contextualized
sentence representations using a BiLSTM equipped with an attention
mechanism. A scene is represented as the weighted sum of the
representations of its sentences.

We also assume that a scene corresponds to a sequence of audio
segments extracted from the movie and a sequence of frames sampled
(with a fixed sampling frequency) from the video. We first
non-linearly project the features of each modality to a lower
dimension and obtain scene-level representations (as the
attention-weighted average of the segments/frames in each scene).  The
two modalities are combined into a joint representation using late
fusion
\cite{frermann-etal-2018-whodunnit,papasarantopoulos-etal-2019-partners}.
The audiovisual similarity $u_{ij}$ (applied in
Eq.~\eqref{eq:similarity}) between scenes $s_i$ and $s_j$ is the dot
product of their fused representations.

\paragraph{Graph Sparsification}
Next, we sparsify graph~$\mathcal{G}$ (or equivalently,
matrix~$\mathcal{A}$) by considering only $k$~neighbors per scene (see Figure~\ref{fig:graph_tp}(d)). Compared to fully
connected graphs, sparse representations are computationally more
efficient and also have shown better classification accuracy
\cite{ozaki-etal-2011-using,Zhu:2005}.  Moreover, we hypothesize that
sparse graphs are crucial for our turning point identification
task. We anticipate the screenplay graph to capture high-level
differences and similarities between movies which would be difficult
to discern when each scene is connected to every other scene.

The most common way to obtain a sparse graph is to construct a $k$-NN
graph by introducing a threshold on the number of nearest
neighbors~$k$
\cite{Szummer:Jaakkola:2002,goldberg-zhu-2006-seeing,niu-etal-2005-word}.
Specifically, we create sparse graph $\mathcal{G}'$ by selecting the
set of neighbors~$\mathcal{P}_i$ for each scene~$s_i$ as follows:
\begin{gather}
    \mathcal{P}_i = \operatorname{argmax}\limits_{j\in[1,N], |\mathcal{P}_i|=k}p_{ij} \label{eq:nei_selection}
  \end{gather}
  where $p_{ij}$ is calculated as in Eq.~(\ref{eq:gumbel_softmax}).
  After removing for each node the neighbors not included in the
  set~$\mathcal{P}_i$, the new graph $\mathcal{G}'$ contains edges
  $|\mathcal{E}'| \ll |\mathcal{E}|$ which are unweighted.

  Instead of a priori deciding on a fixed number of neighbors~$k$ for
  all scenes, which may cause false neighborhood assumptions, we
  treat~$k$ as a parameter to be learned as part of the network which
  computes $p(y_{it} |s_i, \mathcal{D})$, the probability of a scene
  being a TP. Figure \ref{fig:graph_tp}(c) illustrates this
  neighborhood selection module. All connections of $s_i \in
  \mathcal{G}$ serve as input to a non-linear fully-connected layer
  which outputs a probability distribution~$z_i$ over a pre-defined
  set of neighborhood sizes~$[1,C]$. We then select
  $k_i=\operatorname{argmax}_{t\in[1,C]}z_{it}$ as the neighborhood
  size for scene $s_i$ in the sparse graph~$\mathcal{G}'$.

  When deciding on the neighborhood size $k_i$ and the set of
  neighbors $\mathcal{P}_i$ for scene $s_i$, we perform discrete
  choices, which are not differentiable. We address these
  discontinuities in our model by utilizing the Straight-Through
  Estimator \cite{bengio2013estimating}. During the backward pass we
  compute the gradients with the Gumbel-softmax reparametrization
  trick \cite{maddison2017concrete,jang2017categorical}. To better
  approximate the $\operatorname{argmax}$ selections (for $k$ and
  $\mathcal{P}$) during backpropagation, we also add a low temperature
  parameter $\tau=0.1$ \cite{hinton2015distilling} in the
  $\operatorname{softmax}$ function, shown in
  Eq.~\eqref{eq:gumbel_softmax}.

\subsection{Graph Convolutional Networks}
\label{sec:graph-conv-netw}

We rely on graph convolutional networks (GCNs;
\citealt{duvenaud2015convolutional,kearnes2016molecular,kipf2017semi})
to induce embeddings representing graph nodes. Our GCN operates over
the sparsified graph $\mathcal{G}'$ and computes a representation for
the current scene $s_i$ based on the representation of its
neighbors. We only encode information from the scene's
\emph{immediate} neighbors and thus consider one layer of
convolution.\footnote{Performance deteriorates when stacking GCN
  layers.} Moreover, in accordance with \citet{kipf2017semi}, we add a
self-loop to all scenes in $\mathcal{G}'$. This means that the
representation of scene~$s_i$ itself affects the neighborhood
representation~$t_i$:{ \thinmuskip=0mu\medmuskip=0mu\thickmuskip=0mu
 \begin{gather}
    t_{i} = f \left( \frac{1}{|P_i\cup \{s_i\}|}\sum_{j \in P_i \cup
        \{s_i\}}\left( W_{g}c_{j} + b\right) \right)
        \end{gather}
      }where $f(.)$ is a non-linear activation function
      (i.e.,~$\operatorname{ReLU}$), and vectors~$c$ represent the
      \emph{content} of a scene (in relation to the overall screenplay
      and its relative position).

      We encode the screenplay as a sequence $v_1, v_2, ..., v_N$ of textual scene representations
      with a BiLSTM network and obtain contextualized
      representations~$c$ by concatenating the hidden layers of the
      forward~$\overrightarrow{h}$ and backward~$\overleftarrow{h}$
      LSTM ($c = [\overrightarrow{h} ; \overleftarrow{h}]$). In other
      words, graph convolutions are performed on top of LSTM states
      \cite{marcheggiani-titov-2017-encoding}. The one-layer GCN only
      considers information about a scene's immediate neighbors, while
      contextualized scene representations $c$ capture longer-range
      relations between scenes.  Figure~\ref{fig:graph_tp}(e) illustrates
      our GCN and the computation of the neighborhood
      representations~$t$.

      Finally, we concatenate the neighborhood representation $t_i$
      and the content representation $c_i$ to obtain an encoding for
      each scene $s_i$: $[c_i;t_i]$. This vector is fed to a single
      neuron that outputs probabilities $p(y_{it}|s_i,\mathcal{D})$.

\subsection{Model Training}
\label{sec:model-training}

Our description so far has assumed that TP labels are available for
screenplay scenes. However, in practice, such data cannot be easily
sourced (due to the time consuming nature of watching movies, reading
screenplays, and identifying TP locations). The only TP related
dataset we are aware of is TRIPOD \cite{papalampidi2019movie} which
contains TP labels for sentences (not scenes) contained within movie
synopses (not screenplays).  For this reason, we first train a teacher
model which takes as input synopses marked with gold-standard TP
sentences and the corresponding screenplays~$D$ and outputs the
probability $q(y_{it}|s_i,D)$ for scene $s_i$ to convey the meaning of
the~$t^{th}$ TP sentence, where $t \in [1,5]$.

We use the model proposed in \citet{papalampidi2019movie} as teacher
to obtain probability distribution $q(y_t|D)$ over screenplay~$D$.
The TP-specific posterior distributions produced by the teacher model
are used to train our model which only takes scenes as input.
Similarly to knowledge distillation settings
\cite{ba2014deep,hinton2015distilling} we utilize the
KL divergence loss between the teacher posterior distributions
$q(y_t|D)$ and the ones computed by our model~$p(y_t|D)$:
\begin{gather}
 \mathcal{O}_t = \mathcal{D}_{KL}\left(p(y_t|D) \middle\| q(y_t|D) \right), t \in [1,T]
\end{gather}
where $T$ is the number of TPs. We further add a second objective to
the loss function in order to control adjacency matrix~$S$ and hence
the latent graph~$\mathcal{G}'$. Intuitively, we want to assign higher
probabilities for scenes to be neighbors in $\mathcal{G}'$ if they are
also temporally close in the screenplay. For this reason, we add a
focal regularization term~$\mathcal{F}$ to the loss
function. Specifically, we assume a Gaussian distribution $g_i$~over
the screenplay centered around the current scene index~$i$ and try to
keep the probability distribution in matrix~$S$ that corresponds to
candidate neighbors for scene~$s_i$ close to the prior distribution:
$\mathcal{F}_i = \mathcal{D}_{KL}\left(p_i \middle\|
  g_i\right)$.\footnote{We disregard~$\tau$ (see
  Eq.~(\ref{eq:gumbel_softmax})) while recalculating
  probabilities~$p_{ij}$, since we want to directly regulate
  the~$e_{ij}$ values.} The loss function now becomes:
\begin{align}
    \mathcal{L} = \frac{1}{T} \sum_{t=1}^{T}\mathcal{O}_t +
    \lambda \frac{1}{N}\sum_{i=1}^{N}\mathcal{F}_i \label{eq:objective}
    \end{align}
where~$\lambda$ is a hyperparameter.  

\section{Experimental Setup}

\begin{table}[t]
\small
\centering
\begin{tabular}{lrr}
\hline
 & \textbf{Train} & \textbf{Test}  \\ \hline
movies & 84 & 38 \\
scenes & 11,320 & 5,830 \\
TP scenes & 1,260 (SS) & 340 (GS) \\
vocabulary & 37.8k & 28.3k \\ \hline
\multicolumn{3}{c}{\textit{per movie}} \\ \hline
scenes & 133.0 (61.1) & 153.4 (54.0) \\
sentences & 3.0k (0.9) & 2.9k (0.6) \\
tokens & 23.0k (6.6) &  21.5k (4.0) \\
video length (secs) & 6.8k (1.1) & 6.9k (1.3) \\
video frames & 4.2k (3.0) & 3.5k (1.1) \\
audio segments & 12.0k (10.3) & 9.7k (3.1) \\
 \hline
\multicolumn{3}{c}{\textit{per scene}} \\ \hline
sentences & 22.2 (31.5) & 19.0 (24.9) \\ 
tokens & 173.0 (235.0) & 139.9 (177.5) \\ 
sentence tokens & 7.8 (6.0) & 7.4 (6.0) \\
video length (secs) & 88.1 (152.5) & 81.6 (114.8) \\
video frames & 29.2 (37.3) & 23.0 (26.3) \\
audio segments & 82.8 (133.6) & 62.9 (94.0) \\ \hline
\end{tabular}
\caption{Statistics of the augmented TRIPOD dataset; means are shown with
  standard deviation in brackets. SS: silver-standard labels based on \citet{papalampidi2019movie}, GS: gold-standard labels.}
\label{tab:dataset_statistics}
\end{table}

\paragraph{Multimodal TRIPOD}

We performed experiments on the TRIPOD dataset \footnote{\url{https://github.com/ppapalampidi/TRIPOD}}
\cite{papalampidi2019movie} originally used for analyzing the
narrative structure of movies.  We augmented this dataset by
collecting gold-standard annotations for 23~new movies which we added
to the test set. The resulting dataset contains 17,150 scenes from 122 movies, 38
of which have gold-standard scene-level annotations and were used for
evaluation purposes. We also collected the videos and subtitles for
the TRIPOD movies. Table~\ref{tab:dataset_statistics} presents the
dataset statistics.

\vspace*{-.2cm}
\paragraph{Data Preprocessing}
We used the Universal Sentence Encoder (USE;
\citealt{cer2018universal}) to obtain sentence-level
representations. Following previous work \cite{tapaswi2015aligning},
subtitles (and their timestamps on the movie video) were aligned to the dialogue parts of
the screenplay using Dynamic Time Wrapping (DTW;
\citealt{myers1981comparative}). Subsequently, we obtained alignments of
screenplay scenes to video segments. Finally, we segmented the video into scenes and extracted audiovisual features.

For the visual modality, we first sampled one out of every 50 frames
within each scene. However, the length of a scene can vary from a few
seconds to several minutes. For this reason, in cases where the number
of sampled frames became too big for memory, we lowered the sampling
frequency to one frame per~150. We employed ResNeXt-101 \cite{Xie2016}
pre-trained for object recognition on ImageNet \cite{deng2009imagenet}
to extract a visual representation per frame. Similarly, for the audio
modality, we used YAMNet pre-trained on the AudioSet-YouTube corpus
\cite{gemmeke2017audio} for classifying audio segments into 521 audio
classes (e.g.,~tools, music, explosion); for each audio segment
contained in the scene, we extracted features from the penultimate
layer.

\vspace*{-.3cm}
\paragraph{Implementation Details}

Following \cite{papalampidi2019movie}, we select $l=3$ consecutive
scenes to represent each TP in the summary. Moreover, we set the
maximum size of neighbors $C$ that can be selected for a scene in
graph $\mathcal{G}'$ to 6, since we want to create a sparse and
interpretable graph. Experiments with fixed-sized neighborhoods also
showed that performance dropped when considering neighborhoods over 6
scenes.  For training our model we set the hyperparameter $\lambda$ in
Eq.~\eqref{eq:objective} to 10.  We used the Adam algorithm
\cite{kingma2014adam} for optimizing our networks. We chose an LSTM
with 64 neurons for encoding scenes in the screenplay and an identical
one for contextualizing them. We also added a dropout of~0.2. Our
models were developed in PyTorch \cite{paszke2019pytorch} and PyTorch
geometric \cite{Fey/Lenssen/2019}. For analyzing the movie graphs we
used NetworkX \cite{hagberg2008exploring}.

\section{Results}

Our experiments were designed to answer three questions: (1)~Is the
proposed graph-based model better at identifying TPs compared to less
structure-aware variants? (2) To what extent are graphs and multimodal
information helpful? and (3) Are the summaries produced by
automatically identified TPs meaningful?

\vspace*{-.2cm}
\paragraph{Which Model  Identifies TPs Best} 
Table~\ref{tab:main_results} addresses our first question.  We perform
\mbox{5-fold} cross-validation over 38~gold-standard movies to obtain
a test-development split and evaluate model performance in terms of
three metrics: Total Agreement (TA), i.e.,~the percentage of TP scenes
that are correctly identified, Partial Agreement (PA), i.e.,~the
percentage of TP events for which at least one gold-standard scene is
identified, and Distance ($D$), i.e.,~the minimum distance in number
of scenes between the predicted and gold-standard set of scenes for a
given TP, normalized by the screenplay length (see Appendix for a more
detailed definition of the evaluation metrics). We consider TA and PA
as our main evaluation metrics as they measure the percentage of exact
TP matches. However, apart from identifying important events, when
producing a summary it is also important to display events from all
parts of the movie in order to accurately describe its storyline.  For
this reason, we also employ the distance~$D$ metric which quantifies
how well distributed the identified TP events are in the movie. Hence,
a disproprionately large~$D$ suggests that the model fails to even
predict the correct sections of a movie where TPs might be located let
alone the TPs themselves.

\begin{table}[t]
\centering
\small
\begin{tabular}{l|crr}
\hline 
 &  \textbf{TA} $\uparrow$  & \textbf{PA} $\uparrow$  &  \textbf{D} $\downarrow$  \\ \hline
 Random (evenly distributed) & 4.82 & 6.95 & 12.35 \\
 Theory position & 4.41 & 6.32 & 11.03 \\
 Distribution position & 5.59 & 7.37 & 10.74 \\ \hline
 \textrank & 6.18 & 10.00 & 17.77 \\
\begin{tabular}[c]{@{}l@{}} \hspace{1em} + audiovisual \end{tabular} & 6.18 & 10.00 & 18.90 \\
 \scenesum & 4.41 & 7.89 & 16.86  \\
 \begin{tabular}[c]{@{}l@{}} \hspace{1em} + audiovisual \end{tabular} & 6.76 & 11.05 & 18.93 \\ \hline
\begin{tabular}[c]{@{}l@{}} TAM  \end{tabular} & 7.94 & 9.47 & \textbf{9.42} \\
\begin{tabular}[c]{@{}l@{}} \hspace{1em} + audiovisual \end{tabular} & 7.36 & 10.00 & 10.01 \\ \hline
\begin{tabular}[c]{@{}l@{}} \graphtp \end{tabular} & 6.76 & 10.00 & 9.62 \\ 
\begin{tabular}[c]{@{}l@{}} \hspace{1em} + audiovisual \end{tabular} & \textbf{9.12} & \textbf{12.63} & 9.77 \\ 
\hline
\end{tabular}
\caption{Five-fold crossvalidation.  Total Agreement (TA), Partial
  Agreement (PA), and mean distance $D$.}
\label{tab:main_results}
\end{table}

The first block in the table compares our graph-based TP
identification model (henceforth \graphtp) against the following
baselines: a random selection of a sequence of three scenes from five
evenly segmented sections of the movie (reported mean of five runs);
the selection of a sequence of three scenes that lie on the expected
position of each TP event according to screenwriting theory
\cite{Hauge:2017}; and the selection of a sequence of three scenes
based on the position of gold-standard TPs in the synopses of the
TRIPOD training set. The second block includes the performance of two
unsupervised summarization models:
\textrank~\cite{mihalcea2004textrank} with neural input
representations \cite{zheng2019sentence}\footnote{We also experimented
  with directed \textrank~\cite{zheng2019sentence}, but these results
  were poor and are omitted for the sake of brevity.} and
\scenesum~\cite{papalampidi2020screenplay,gorinski-lapata-2015-movie},
a variant of \textrank~that takes the characters participating in each
scene into account.  Finally, we report results for the Topic-Aware
Model (TAM; \citealt{papalampidi2020screenplay}); TAM is
sequence-based supervised model which employs a sliding context window
and computes the similarity between sequential contexts. We discuss
implementation details for comparison systems in the Appendix.  We
also report the performance of a multimodal variant for all comparison
systems (+audiovisual).  For the unsupervised models, we add
scene-level features as extra weights to the pairwise similarity
calculation between scenes similarly to \graphtp. For TAM, we add
audiovisual information via early fusion; we concatenate the
scene-level vectors from all modalities (i.e.,~text, vision, audio).

The unsupervised summarization models (\textrank, \scenesum) have
competitive performance in terms of TA and PA, but significantly
higher average distance~D.  This suggests that they do not select
events from all parts of a story but favor specific sections. For the
supervised models (TAM and \graphtp), the average~D is in general
lower, which means that they are able to adapt to the positional bias
and select events from all parts of a movie. Moreover, both models
seem to benefit from multimodal information (see TA and PA metrics).
Finally, \graphtp~seems to perform best, by correctly identifying a
higher number of gold-standard TP events (based on both TA and PA
metrics), whereas~D is comparable for TAM and \graphtp.

\begin{table}[t]
\centering
\small
\begin{tabular}{l|crr}
\hline 
 & \begin{tabular}[c]{@{}c@{}} \textbf{TA} $\uparrow$ \end{tabular} & \begin{tabular}[c]{@{}c@{}} \textbf{PA} $\uparrow$ \end{tabular} & \begin{tabular}[c]{@{}c@{}} \textbf{D} $\downarrow$ \end{tabular} \\ \hline
\begin{tabular}[c]{@{}l@{}} Fully connected graph \end{tabular} & 5.00 & 7.37 & 9.73  \\ 
\begin{tabular}[c]{@{}l@{}} Content  \end{tabular} & 5.59 & 7.37 & 9.98 \\
\begin{tabular}[c]{@{}l@{}} Neighborhood \& position \end{tabular} & \textbf{9.71} & \textbf{13.16} & 10.98  \\  
\begin{tabular}[c]{@{}l@{}} \graphtp \end{tabular} & 6.76 & 10.00 & \textbf{9.62}  \\ 
\begin{tabular}[c]{@{}l@{}} \hspace{1em} + vision \end{tabular} & 6.18 & 7.89 & 9.84 \\ 
\begin{tabular}[c]{@{}l@{}} \hspace{1em} + audio \end{tabular}  & 7.06 & 8.95 & 10.38 \\
\begin{tabular}[c]{@{}l@{}} \hspace{1em} + audiovisual \end{tabular} & 9.12 & 12.63 & 9.77 \\ 
\hline
\end{tabular}
\caption{\graphtp variants. Total Agreement (TA), Partial Agreement
  (PA), and mean distance~D.}
\label{tab:ablation}
\end{table}

\begin{table}[t]
\centering
  \begin{small}
\begin{tabular}{lrrrr} \hline
&\scenesum & TAM & \graphtp  & Gold \\ \hline
 TP1 & 28 &  28 & \textbf{64}& {66}\\
 TP2 & 36 & 54 &\textbf{64}& 62 \\
 TP3 & 38 &  18 & \textbf{44}& {54}\\
 TP4 & 26 & 34 & \textbf{52}& {56}\\
 TP5 & 8 & 16 & \textbf{24} &{48}\\
 Mean  &27 & 30 & \textbf{50}  & {57}\\
 Rating  &2.63&2.68& \textbf{3.02}& {3.58}\\ \hline
\end{tabular}
\caption{\label{tab:human_evaluation} Human evaluation; proportion of
  TPs found in video summaries (shown as percentages) and average
  ratings attributed to each system (ratings vary from 1 to 5, with 5
  being best). All pairwise differences are significant ($p<0.05$,
  using a $\chi^{2}$ test).}
\end{small}
\end{table}

\vspace*{-.2cm}
\paragraph{Which Information Matters}
Table~\ref{tab:ablation} answers our second question by presenting an
ablation study on \graphtp. We observe that the performance of a
similar model which uses a fully-connected graph drops across
metrics. This is also the case when we do not take into account a
graph or any other form of interaction between scenes (i.e.,~only
content). We also test the model's performance when we remove the
content representation and keep only the neighborhood interactions
together with a vector that simply encodes the position of a scene in
the screenplay (as a one-hot vector). This model may not be able to
adapt to the positional bias as well (higher D), but is able to
predict TPs with higher accuracy (high TA and PA). Finally, we find
that audio and visual information boost model performance in
combination but not individually.

\vspace*{-.2cm}
\paragraph{How Good Are the Summaries} We now answer our last question
by evaluating the video summaries produced by our model. We conducted
a human evaluation experiment using the videos created for 10 movies
of the test set. We produced summaries based on the gold scene-level
annotations (Gold), \scenesum, TAM, and \graphtp (see Appendix). For all systems we used model variants which consider audiovisual information except for
\scenesum. Inclusion of audiovisual information for this model yields
overly long summaries (in the excess of 30 minutes) for most
movies. All other systems produce 15 minutes long summaries on
average.

\begin{figure}
    \centering
    \begin{tikzpicture}[scale=.9]
      \begin{axis}[%
        height=5cm,
        width=9cm, 
        ymax=.9,
        ymin=0.3,
        axis y line*=left,
        axis x line*=bottom,
                symbolic x coords={TP1,TP2,TP3,TP4,TP5},
        x tick label style={rotate=0},
                xticklabel={\tick},
        legend style={at={(0.5,-0.17)},anchor=north,
        legend cell align=left, 
        legend columns=2, /tikz/column 2/.style={
                column sep=5pt,
            },
        }
      ]
      \addplot plot coordinates {
      (TP1, 0.7595683675393063)
      (TP2, 0.7566194895208482)
      (TP3, 0.5930124569554308)
      (TP4, 0.5799460551903782)
      (TP5, 0.5974631384559111)
      };
      \addlegendentry{Comedy/Romance}
      \addplot plot coordinates {
      (TP1, 0.36828863175525905)
      (TP2, 0.46757092696877206)
      (TP3, 0.3895776754566795)
      (TP4, 0.4497085358848819)
      (TP5, 0.5316365896266323)
      };
      \addlegendentry{Thriller/Mystery}
      \addplot plot coordinates {
      (TP1, 0.7823899355464081)
      (TP2, 0.7919933698527087)
      (TP3, 0.8677384337463844)
      (TP4, 0.8572568106438652)
      (TP5, 0.7990329420595408)
      };
      \addlegendentry{Action}
      \addplot plot coordinates {
      (TP1, 0.6743128769997379)
      (TP2, 0.6619415168274334)
      (TP3, 0.6100698938382695)
      (TP4, 0.479330137691664)
      (TP5, 0.6843946509387108)
      };
      \addlegendentry{Drama/Other}
      \end{axis}
    \end{tikzpicture}%
        \caption{Average node connectivity per TP across movie genres (\graphtp
      (+audiovisual), test set.)}
          \label{fig:connectivity_per_tp}
  \end{figure}
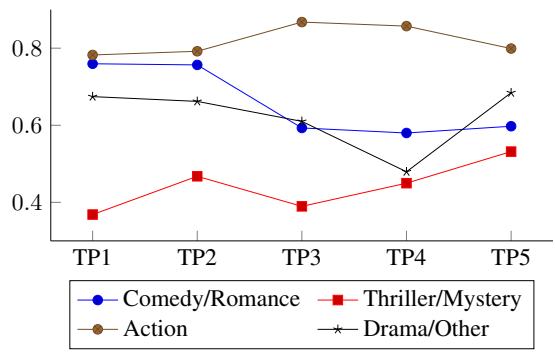

  Our study was conducted on Amazon Mechanical Turk
  (AMT). Crowdworkers first read a short summary of the movie
  (i.e.,~an abridged version of the Wikipedia plot
  synopsis). Subsequently, they were asked to watch a video summary
  and answer five questions each pertaining to a specific TP event
  (described in the textual summary).  AMT workers answered with `Yes'
  if they were certain it was present in the video, `No' if the event
  was absent, and `Unsure' otherwise. Finally, we asked AMT workers to
  provide an overall rating from 1 to 5, with 5 being the most
  informative summary.  Workers were asked to take into account the
  questions answered previously, but also consider the overall quality
  of the summary (i.e., how compressed it was, whether it contained
  redundant events, and the overall information provided).  We asked 5
  different workers to evaluate each movie summary.  Instructions and
  example questions are in the Appendix.

  Table~\ref{tab:human_evaluation} shows the proportion of `Yes'
  answers (per TP and overall mean) and the average system
  rating.\footnote{We omit 'Unsure' from
    Table~\ref{tab:human_evaluation}, since it only accounts for~4.1\%
    of the answers.} Perhaps unsurprisingly gold summaries are the
  most informative. Some key events might still be absent due to
  errors in the automatic alignment between the screenplay scenes and
  the video. \graphtp~is the second best system overall (and across
  TPs), while \scenesum~and TAM have similar ratings.  \graphtp
  manages to create more informative and diverse summaries, presenting
  important events from all parts of the story.

\begin{figure}[t]
    \tiny
    \begin{tabular}{|c|c|} \hline
        {{\small Die Hard}} & {{\small American Beauty}}  \\  
         \includegraphics[width=0.44\columnwidth,page=1]{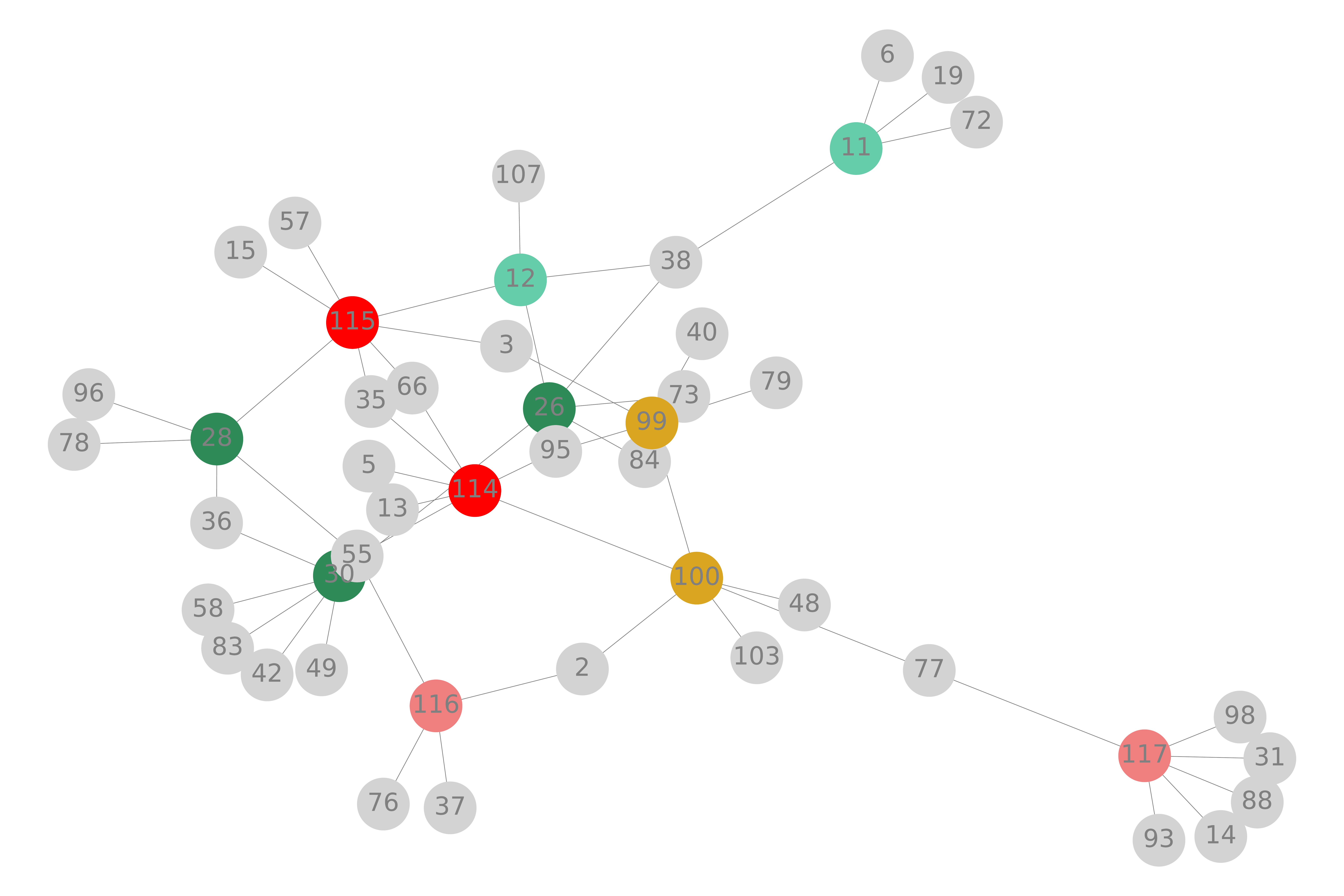} & 
        \includegraphics[width=0.44\columnwidth,page=1]{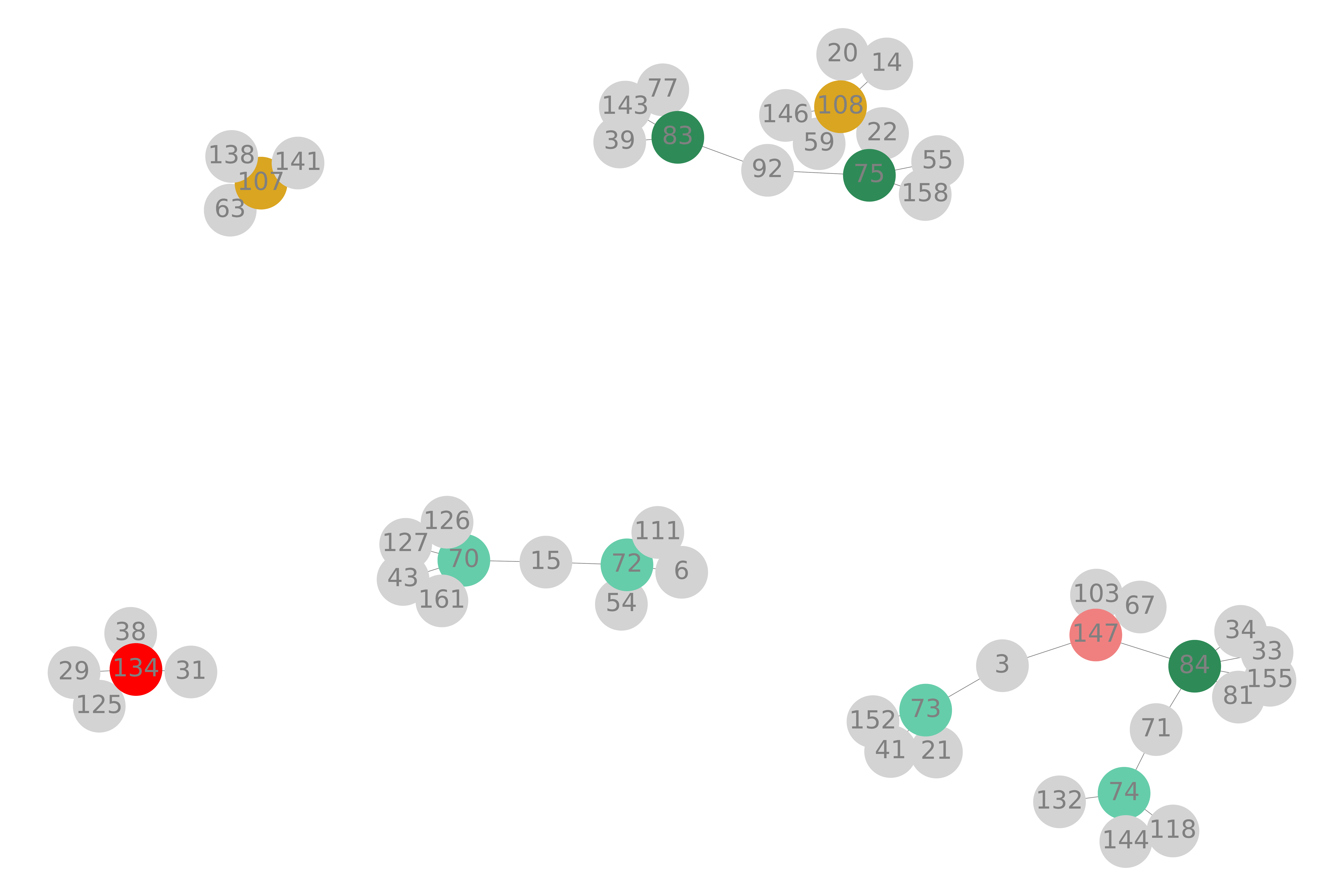}\\\hline
    \end{tabular}
 \caption{Examples of graphs produced by \graphtp (+audiovisual) for
   two movies from the test set. Nodes  (in color) are scenes which
   act as TPs and their immediate neighbors (in gray). 
   \label{fig:graph_examples}}
\end{figure}

\vspace*{-.2cm}
  \paragraph{What Do the Graphs Mean} We further analyzed the graphs
  induced by our model, in particular their connectivity.
  Figure~\ref{fig:connectivity_per_tp} shows the average node
  connectivity per TP (i.e.,~minimum number of nodes that need to be
  removed to separate the remaining nodes into isolated subgraphs) for
  the movies in the test set. For this analysis, graphs were pruned to
  nodes which act as TPs and their immediate neighbors and movies were
  grouped in four broad genre categories ({comedy/romance},
  {thriller/mystery}, {action} and {drama/other}).  We find that
  thrillers and mysteries correspond to more disconnected graphs
  followed by dramas, while comedies, romance and especially action
  movies display more connected graphs. This is intuitive, since
  comedies and action movies tend to follow predictable storylines,
  while thrillers often contain surprising events which break
  screenwriting conventions. 
Moreover, for comedies and dramas the introductory events (i.e., first two TPs) are the central ones in the graph and connectivity decreases as the story unfolds and unexpected events take place. We see the opposite trend for thrillers and action movies. Initial events present lower connectivity, while the last ones are now central when crucial information is revealed justifying earlier actions (e.g., see the last two TPs which correspond to `major setback' and `climax').  
A similar picture emerges when visualizing the graphs (see
Figure~\ref{fig:graph_examples}).  ``Die Hard'', a conventional action movie, has
a clear storyline and seems more connected, while ''American Beauty'',
a drama with several flashbacks, contains several disconnected subgraphs
(see Appendix for more illustrations).

\section{Conclusions}

In this paper we demonstrate that TP identification can be used directly for summarizing movies. We propose \graphtp, a model that operates over sparse graphs relying on multimodal information. Summaries created by \graphtp are preferred by humans in contrast to general summarization algorithms and sequence-based models. 
In the future, we will explore ways to further exploit the graph structure and definitions of TPs in order to produce personalized summaries. 

\section*{Acknowledgments}
We thank the anonymous reviewers for their feedback.  
We gratefully
acknowledge the support of the European Research Council (Lapata;
award 681760, ``Translating Multiple Modalities into Text'') and of
the Leverhulme Trust (Keller; award IAF-2017-019).

\bibliography{aaai2020}

\begin{thebibliography}{68}
\providecommand{\natexlab}[1]{#1}
\providecommand{\url}[1]{\texttt{#1}}
\providecommand{\urlprefix}{URL }
\expandafter\ifx\csname urlstyle\endcsname\relax
  \providecommand{\doi}[1]{doi:\discretionary{}{}{}#1}\else
  \providecommand{\doi}{doi:\discretionary{}{}{}\begingroup
  \urlstyle{rm}\Url}\fi

\bibitem[{Ba and Caruana(2014)}]{ba2014deep}
Ba, J.; and Caruana, R. 2014.
\newblock Do Deep Nets Really Need to be Deep?
\newblock In \emph{Proceedings of the Advances in NeurIPS}.

\bibitem[{Bengio, L{\'e}onard, and Courville(2013)}]{bengio2013estimating}
Bengio, Y.; L{\'e}onard, N.; and Courville, A. 2013.
\newblock Estimating or propagating gradients through stochastic neurons for
  conditional computation.
\newblock \emph{arXiv preprint arXiv:1308.3432} .

\bibitem[{Black and Wilensky(1979)}]{black1979evaluation}
Black, J.~B.; and Wilensky, R. 1979.
\newblock An evaluation of story grammars.
\newblock \emph{Cognitive science} 3(3): 213--229.

\bibitem[{Cer et~al.(2018{\natexlab{a}})Cer, Yang, Kong, Hua, Limtiaco, John,
  Constant, Guajardo-Cespedes, Yuan, Tar et~al.}]{cer2018universal}
Cer, D.; Yang, Y.; Kong, S.-y.; Hua, N.; Limtiaco, N.; John, R.~S.; Constant,
  N.; Guajardo-Cespedes, M.; Yuan, S.; Tar, C.; et~al. 2018{\natexlab{a}}.
\newblock Universal sentence encoder.
\newblock \emph{arXiv preprint arXiv:1803.11175} .

\bibitem[{Cer et~al.(2018{\natexlab{b}})Cer, Yang, Kong, Hua, Limtiaco,
  St.~John, Constant, Guajardo-Cespedes, Yuan, Tar, Strope, and
  Kurzweil}]{cer-etal-2018-universal}
Cer, D.; Yang, Y.; Kong, S.-y.; Hua, N.; Limtiaco, N.; St.~John, R.; Constant,
  N.; Guajardo-Cespedes, M.; Yuan, S.; Tar, C.; Strope, B.; and Kurzweil, R.
  2018{\natexlab{b}}.
\newblock Universal Sentence Encoder for {E}nglish.
\newblock In \emph{Proceedings of the 2018 Conference on EMNLP: System
  Demonstrations}.

\bibitem[{Chambers and Jurafsky(2009)}]{chambers-jurafsky-2009-unsupervised}
Chambers, N.; and Jurafsky, D. 2009.
\newblock Unsupervised Learning of Narrative Schemas and their Participants.
\newblock In \emph{Proceedings of the 47th Annual Meeting of ACL and the 4th
  IJCNLP}.

\bibitem[{Cutting(2016)}]{cutting2016narrative}
Cutting, J.~E. 2016.
\newblock Narrative theory and the dynamics of popular movies.
\newblock \emph{Psychonomic bulletin \& review} 23(6).

\bibitem[{Deng et~al.(2009)Deng, Dong, Socher, Li, Li, and
  Fei-Fei}]{deng2009imagenet}
Deng, J.; Dong, W.; Socher, R.; Li, L.-J.; Li, K.; and Fei-Fei, L. 2009.
\newblock Imagenet: A large-scale hierarchical image database.
\newblock In \emph{2009 IEEE conference on CVPR}, 248--255. Ieee.

\bibitem[{Do, Tran, and Tran(2018)}]{do2018movie}
Do, T. T.~H.; Tran, Q. H.~B.; and Tran, Q.~D. 2018.
\newblock Movie indexing and summarization using social network techniques.
\newblock \emph{Vietnam Journal of Computer Science} 5(2): 157--164.

\bibitem[{Duvenaud et~al.(2015)Duvenaud, Maclaurin, Iparraguirre, Bombarell,
  Hirzel, Aspuru-Guzik, and Adams}]{duvenaud2015convolutional}
Duvenaud, D.~K.; Maclaurin, D.; Iparraguirre, J.; Bombarell, R.; Hirzel, T.;
  Aspuru-Guzik, A.; and Adams, R.~P. 2015.
\newblock Convolutional networks on graphs for learning molecular fingerprints.
\newblock In \emph{Advances in NeurIPS}, 2224--2232.

\bibitem[{Fey and Lenssen(2019)}]{Fey/Lenssen/2019}
Fey, M.; and Lenssen, J.~E. 2019.
\newblock Fast Graph Representation Learning with {PyTorch Geometric}.
\newblock In \emph{ICLR Workshop on Representation Learning on Graphs and
  Manifolds}.

\bibitem[{Finlayson(2012)}]{Finlayson:2012}
Finlayson, M.~A. 2012.
\newblock \emph{Learning Narrative Structure from Annotated Folktales}.
\newblock Ph.D. thesis, MIT.

\bibitem[{Frermann, Cohen, and Lapata(2018)}]{frermann-etal-2018-whodunnit}
Frermann, L.; Cohen, S.~B.; and Lapata, M. 2018.
\newblock Whodunnit? Crime Drama as a Case for Natural Language Understanding.
\newblock \emph{TACL} 6.

\bibitem[{Fujii, Kitaoka, and Nakagawa(2007)}]{Fujii2007AutomaticEO}
Fujii, Y.; Kitaoka, N.; and Nakagawa, S. 2007.
\newblock Automatic extraction of cue phrases for important sentences in
  lecture speech and automatic lecture speech summarization.
\newblock In \emph{INTERSPEECH}.

\bibitem[{Gemmeke et~al.(2017)Gemmeke, Ellis, Freedman, Jansen, Lawrence,
  Moore, Plakal, and Ritter}]{gemmeke2017audio}
Gemmeke, J.~F.; Ellis, D.~P.; Freedman, D.; Jansen, A.; Lawrence, W.; Moore,
  R.~C.; Plakal, M.; and Ritter, M. 2017.
\newblock Audio set: An ontology and human-labeled dataset for audio events.
\newblock In \emph{2017 IEEE ICASSP}, 776--780. IEEE.

\bibitem[{Goldberg and Zhu(2006)}]{goldberg-zhu-2006-seeing}
Goldberg, A.; and Zhu, X. 2006.
\newblock Seeing stars when there aren{'}t many stars: Graph-based
  semi-supervised learning for sentiment categorization.
\newblock In \emph{Proceedings of {T}ext{G}raphs: the First Workshop on Graph
  Based Methods for NLP}.

\bibitem[{Gorinski and Lapata(2015)}]{gorinski-lapata-2015-movie}
Gorinski, P.~J.; and Lapata, M. 2015.
\newblock Movie Script Summarization as Graph-based Scene Extraction.
\newblock In \emph{Proceedings of the 2015 Conference of NAACL-HLT}.

\bibitem[{Goyal, Riloff, and
  Daum{\'e}~III(2010)}]{goyal-etal-2010-automatically}
Goyal, A.; Riloff, E.; and Daum{\'e}~III, H. 2010.
\newblock Automatically producing plot unit representations for narrative text.
\newblock In \emph{Proceedings of the 2010 Conference on EMNLP}, 77--86.

\bibitem[{Grusky, Naaman, and Artzi(2018)}]{newsroom-naacl18}
Grusky, M.; Naaman, M.; and Artzi, Y. 2018.
\newblock {NEWSROOM: A} Dataset of 1.3 Million Summaries with Diverse
  Extractive Strategies.
\newblock In \emph{Proceedings of the 16th Annual Conference of NAACL-HLT}.

\bibitem[{Hagberg, Swart, and S~Chult(2008)}]{hagberg2008exploring}
Hagberg, A.; Swart, P.; and S~Chult, D. 2008.
\newblock Exploring network structure, dynamics, and function using NetworkX.
\newblock Technical report.

\bibitem[{Hauge(2017)}]{Hauge:2017}
Hauge, M. 2017.
\newblock \emph{Storytelling Made Easy: Persuade and Transform Your Audiences,
  Buyers, and Clients -- Simply, Quickly, and Profitably}.
\newblock Indie Books International.

\bibitem[{Hinton, Vinyals, and Dean(2015)}]{hinton2015distilling}
Hinton, G.; Vinyals, O.; and Dean, J. 2015.
\newblock Distilling the knowledge in a neural network.
\newblock \emph{arXiv preprint arXiv:1503.02531} .

\bibitem[{Jang, Gu, and Poole(2017)}]{jang2017categorical}
Jang, E.; Gu, S.; and Poole, B. 2017.
\newblock Categorical Reparametrization with Gumble-Softmax.
\newblock In \emph{ICLR}.

\bibitem[{Jung et~al.(2019)Jung, Kang, Mentch, and Hovy}]{jung2019earlier}
Jung, T.; Kang, D.; Mentch, L.; and Hovy, E. 2019.
\newblock Earlier Isn't Always Better: Sub-aspect Analysis on Corpus and System
  Biases in Summarization.
\newblock \emph{arXiv preprint arXiv:1908.11723} .

\bibitem[{Kazantseva and
  Szpakowicz(2010)}]{kazantseva-szpakowicz-2010-summarizing}
Kazantseva, A.; and Szpakowicz, S. 2010.
\newblock Summarizing Short Stories.
\newblock \emph{Computational Linguistics} 36(1).

\bibitem[{Kearnes et~al.(2016)Kearnes, McCloskey, Berndl, Pande, and
  Riley}]{kearnes2016molecular}
Kearnes, S.; McCloskey, K.; Berndl, M.; Pande, V.; and Riley, P. 2016.
\newblock Molecular graph convolutions: moving beyond fingerprints.
\newblock \emph{Journal of computer-aided molecular design} 30(8): 595--608.

\bibitem[{Kingma and Ba(2014)}]{kingma2014adam}
Kingma, D.~P.; and Ba, J. 2014.
\newblock Adam: A method for stochastic optimization.
\newblock \emph{arXiv preprint arXiv:1412.6980} .

\bibitem[{Kipf and Welling(2017)}]{kipf2017semi}
Kipf, T.~N.; and Welling, M. 2017.
\newblock Semi-Supervised Classification with Graph Convolutional Networks.
\newblock In \emph{ICLR}.

\bibitem[{Lehnert(1981)}]{Lehnert:1981}
Lehnert, W.~G. 1981.
\newblock Plot Units and Narrative Summarization.
\newblock \emph{Cognitive Science} 5(4): 293--331.

\bibitem[{Liu and Lapata(2019)}]{liu-lapata-2019-hierarchical}
Liu, Y.; and Lapata, M. 2019.
\newblock Hierarchical Transformers for Multi-Document Summarization.
\newblock In \emph{Proceedings of the 57th Annual Meeting of ACL}.

\bibitem[{Lohnert, Black, and Reiser(1981)}]{lohnert1981summarizing}
Lohnert, W.~G.; Black, J.~B.; and Reiser, B.~J. 1981.
\newblock Summarizing narratives.
\newblock In \emph{Proceedings of the 7th international joint conference on
  Artificial intelligence-Volume 1}, 184--189.

\bibitem[{Maddison, Mnih, and Teh(2017)}]{maddison2017concrete}
Maddison, C.~J.; Mnih, A.; and Teh, Y.~W. 2017.
\newblock The Concrete Distribution: {A} Continuous Relaxation of Discrete
  Random Variables.
\newblock In \emph{5th ICLR}. OpenReview.net.

\bibitem[{Mani(2012)}]{Mani:2012}
Mani, I. 2012.
\newblock \emph{Computational Modeling of Narative}.
\newblock Synthesis Lectures on Human Language Technologies. Morgan and
  Claypool Publishers.

\bibitem[{Marcheggiani and Titov(2017)}]{marcheggiani-titov-2017-encoding}
Marcheggiani, D.; and Titov, I. 2017.
\newblock Encoding Sentences with Graph Convolutional Networks for Semantic
  Role Labeling.
\newblock In \emph{Proceedings of the 2017 Conference on EMNLP}.

\bibitem[{McInnes et~al.(2018)McInnes, Healy, Saul, and
  Gro{\ss}berger}]{mcinnes2018umap}
McInnes, L.; Healy, J.; Saul, N.; and Gro{\ss}berger, L. 2018.
\newblock UMAP: Uniform Manifold Approximation and Projection.
\newblock \emph{Journal of Open Source Software} 3(29): 861.

\bibitem[{McIntyre and Lapata(2010)}]{mcintyre-lapata-2010-plot}
McIntyre, N.; and Lapata, M. 2010.
\newblock Plot Induction and Evolutionary Search for Story Generation.
\newblock In \emph{Proceedings of the 48th Annual Meeting of ACL}.

\bibitem[{Mihalcea and Ceylan(2007)}]{mihalcea-ceylan-2007-explorations}
Mihalcea, R.; and Ceylan, H. 2007.
\newblock Explorations in Automatic Book Summarization.
\newblock In \emph{Proceedings of the 2007 Joint Conference on EMNLP and
  CoNLL}.

\bibitem[{Mihalcea and Tarau(2004)}]{mihalcea2004textrank}
Mihalcea, R.; and Tarau, P. 2004.
\newblock Textrank: Bringing order into text.
\newblock In \emph{Proceedings of the 2004 conference of EMNLP}.

\bibitem[{Murray et~al.(2007)Murray, Hsueh, Tucker, Kilgour, Carletta, Moore,
  and Renals}]{murray2007automatic}
Murray, G.; Hsueh, P.-Y.; Tucker, S.; Kilgour, J.; Carletta, J.; Moore, J.~D.;
  and Renals, S. 2007.
\newblock Automatic segmentation and summarization of meeting speech.
\newblock In \emph{Proceedings of NAACL-HLT}, 9--10.

\bibitem[{Myers and Rabiner(1981)}]{myers1981comparative}
Myers, C.~S.; and Rabiner, L.~R. 1981.
\newblock A comparative study of several dynamic time-warping algorithms for
  connected-word recognition.
\newblock \emph{Bell System Technical Journal} 60(7).

\bibitem[{Narayan, Cohen, and Lapata(2018)}]{narayan-etal-2018-dont}
Narayan, S.; Cohen, S.~B.; and Lapata, M. 2018.
\newblock Don{'}t Give Me the Details, Just the Summary! Topic-Aware
  Convolutional Neural Networks for Extreme Summarization.
\newblock In \emph{Proceedings of the 2018 Conference on EMNLP}.

\bibitem[{Niu, Ji, and Tan(2005)}]{niu-etal-2005-word}
Niu, Z.-Y.; Ji, D.-H.; and Tan, C.~L. 2005.
\newblock Word Sense Disambiguation Using Label Propagation Based
  Semi-Supervised Learning.
\newblock In \emph{Proceedings of the 43rd Annual Meeting of ACL}.

\bibitem[{Ozaki et~al.(2011)Ozaki, Shimbo, Komachi, and
  Matsumoto}]{ozaki-etal-2011-using}
Ozaki, K.; Shimbo, M.; Komachi, M.; and Matsumoto, Y. 2011.
\newblock Using the mutual k-nearest neighbor graphs for semi-supervised
  classification on natural language data.
\newblock In \emph{Proceedings of the fifteenth conference of CoNLL}, 154--162.

\bibitem[{Papalampidi et~al.(2020)Papalampidi, Keller, Frermann, and
  Lapata}]{papalampidi2020screenplay}
Papalampidi, P.; Keller, F.; Frermann, L.; and Lapata, M. 2020.
\newblock Screenplay Summarization Using Latent Narrative Structure.
\newblock In \emph{Proceedings of the 58th Annual Meeting of ACL}.

\bibitem[{Papalampidi, Keller, and Lapata(2019)}]{papalampidi2019movie}
Papalampidi, P.; Keller, F.; and Lapata, M. 2019.
\newblock Movie Plot Analysis via Turning Point Identification.
\newblock In \emph{Proceedings of the 2019 Conference on EMNLP and the 9th
  IJCNLP}.

\bibitem[{Papasarantopoulos et~al.(2019)Papasarantopoulos, Frermann, Lapata,
  and Cohen}]{papasarantopoulos-etal-2019-partners}
Papasarantopoulos, N.; Frermann, L.; Lapata, M.; and Cohen, S.~B. 2019.
\newblock Partners in Crime: Multi-view Sequential Inference for Movie
  Understanding.
\newblock In \emph{Proceedings of the 2019 Conference on EMNLP and the 9th
  IJCNLP}.

\bibitem[{Paszke et~al.(2019)Paszke, Gross, Massa, Lerer, Bradbury, Chanan,
  Killeen, Lin, Gimelshein, Antiga et~al.}]{paszke2019pytorch}
Paszke, A.; Gross, S.; Massa, F.; Lerer, A.; Bradbury, J.; Chanan, G.; Killeen,
  T.; Lin, Z.; Gimelshein, N.; Antiga, L.; et~al. 2019.
\newblock PyTorch: An imperative style, high-performance deep learning library.
\newblock In \emph{Advances in NeurIPS}, 8024--8035.

\bibitem[{Propp(1968)}]{Propp:1968}
Propp, V.~I. 1968.
\newblock \emph{Morphology of the Folktale}.
\newblock University of Texas.

\bibitem[{Richards, Finlayson, and Winston(2009)}]{Richards:ea:2009}
Richards, W.; Finlayson, M.~A.; and Winston, P.~H. 2009.
\newblock Advancing Computational Models of Narrative.
\newblock Technical Report 63:2009, MIT Computer Science and Atrificial
  Intelligence Laboratory.

\bibitem[{Rohrbach et~al.(2015)Rohrbach, Rohrbach, Tandon, and
  Schiele}]{rohrbach2015dataset}
Rohrbach, A.; Rohrbach, M.; Tandon, N.; and Schiele, B. 2015.
\newblock A dataset for movie description.
\newblock In \emph{Proceedings of the IEEE conference on CVPR}.

\bibitem[{Rumelhart(1980)}]{Rumelhart:1980}
Rumelhart, D.~E. 1980.
\newblock On evaluating story grammars.
\newblock \emph{Cognitive Science} 4(3): 313--316.

\bibitem[{Schank and Abelson(1975)}]{Schank:Abelson:1975}
Schank, R.~C.; and Abelson, R.~P. 1975.
\newblock Scripts, Plans, and Knowledge.
\newblock In \emph{Proceedings of the 4th International Joint Conference on
  Artificial Intelligence}, 151--157. Tblisi, USSR.

\bibitem[{Srivastava, Chaturvedi, and Mitchell(2016)}]{Srivastava:ea:2016}
Srivastava, S.; Chaturvedi, S.; and Mitchell, T. 2016.
\newblock Inferring Interpersonal Relations in Narrative Summaries.
\newblock In \emph{Proceedings of the 13th AAAI Conference}.

\bibitem[{Syed et~al.(2018)Syed, V{\"o}lske, Potthast, Lipka, Stein, and
  Sch{\"u}tze}]{syed2018task}
Syed, S.; V{\"o}lske, M.; Potthast, M.; Lipka, N.; Stein, B.; and Sch{\"u}tze,
  H. 2018.
\newblock Task Proposal: The TL; DR Challenge.
\newblock In \emph{Proceedings of the 11th International Conference on Natural
  Language Generation}, 318--321.

\bibitem[{Szummer and Jaakkola(2003)}]{Szummer:Jaakkola:2002}
Szummer, M.; and Jaakkola, T.~S. 2003.
\newblock Information Regularization with Partially Labeled Data.
\newblock In Becker, S.; Thrun, S.; and Obermayer, K., eds., \emph{Advances in
  NeurIPS 15}.

\bibitem[{Tapaswi, B{\"a}uml, and Stiefelhagen(2015)}]{tapaswi2015aligning}
Tapaswi, M.; B{\"a}uml, M.; and Stiefelhagen, R. 2015.
\newblock Aligning plot synopses to videos for story-based retrieval.
\newblock \emph{International Journal of Multimedia Information Retrieval}
  4(1).

\bibitem[{Tapaswi, Bauml, and Stiefelhagen(2015)}]{tapaswi2015book2movie}
Tapaswi, M.; Bauml, M.; and Stiefelhagen, R. 2015.
\newblock Book2movie: Aligning video scenes with book chapters.
\newblock In \emph{Proceedings of the IEEE Conference on CVPR}, 1827--1835.

\bibitem[{Tapaswi et~al.(2016)Tapaswi, Zhu, Stiefelhagen, Torralba, Urtasun,
  and Fidler}]{tapaswi2016movieqa}
Tapaswi, M.; Zhu, Y.; Stiefelhagen, R.; Torralba, A.; Urtasun, R.; and Fidler,
  S. 2016.
\newblock Movieqa: Understanding stories in movies through question-answering.
\newblock In \emph{Proceedings of the IEEE conference on CVPR}.

\bibitem[{Teufel and Moens(2002)}]{teufel-moens-2002-articles}
Teufel, S.; and Moens, M. 2002.
\newblock Articles Summarizing Scientific Articles: Experiments with Relevance
  and Rhetorical Status.
\newblock \emph{Computational Linguistics} 28(4).

\bibitem[{Thompson(1999)}]{thompson1999storytelling}
Thompson, K. 1999.
\newblock \emph{Storytelling in the new Hollywood: Understanding classical
  narrative technique}.
\newblock Harvard University Press.

\bibitem[{Tran et~al.(2017)Tran, Hwang, Lee, and Jung}]{tran2017exploiting}
Tran, Q.~D.; Hwang, D.; Lee, O.-J.; and Jung, J.~E. 2017.
\newblock Exploiting character networks for movie summarization.
\newblock \emph{Multimedia Tools and Applications} 76(8): 10357--10369.

\bibitem[{Tsoneva, Barbieri, and Weda(2007)}]{tsoneva2007automated}
Tsoneva, T.; Barbieri, M.; and Weda, H. 2007.
\newblock Automated summarization of narrative video on a semantic level.
\newblock In \emph{ICSC}.

\bibitem[{Valls-Vargas, Zhu, and Ontanon(2014)}]{Vargas:ea:2014}
Valls-Vargas, J.; Zhu, J.; and Ontanon, S. 2014.
\newblock Toward automatic role identification in unannotated folk tales.
\newblock In \emph{Proceedings of the 10th AAAI Conference}, 188--194.

\bibitem[{Wang et~al.(2020)Wang, Liu, Zheng, Qiu, and
  Huang}]{wang-etal-2020-heterogeneous}
Wang, D.; Liu, P.; Zheng, Y.; Qiu, X.; and Huang, X. 2020.
\newblock Heterogeneous Graph Neural Networks for Extractive Document
  Summarization.
\newblock In \emph{Proceedings of the 58th Annual Meeting of ACL}.

\bibitem[{Xie et~al.(2016)Xie, Girshick, Dollár, Tu, and He}]{Xie2016}
Xie, S.; Girshick, R.; Dollár, P.; Tu, Z.; and He, K. 2016.
\newblock Aggregated Residual Transformations for Deep Neural Networks.
\newblock \emph{arXiv preprint arXiv:1611.05431} .

\bibitem[{Xiong et~al.(2019)Xiong, Huang, Guo, Zhou, Zhou, and
  Lin}]{xiong2019graph}
Xiong, Y.; Huang, Q.; Guo, L.; Zhou, H.; Zhou, B.; and Lin, D. 2019.
\newblock A Graph-Based Framework to Bridge Movies and Synopses.
\newblock In \emph{Proceedings of the IEEE ICCV}, 4592--4601.

\bibitem[{Zheng and Lapata(2019)}]{zheng2019sentence}
Zheng, H.; and Lapata, M. 2019.
\newblock Sentence Centrality Revisited for Unsupervised Summarization.
\newblock \emph{arXiv preprint arXiv:1906.03508} .

\bibitem[{Zhu(2005)}]{Zhu:2005}
Zhu, X. 2005.
\newblock \emph{Semi-supervised Learning with Graphs}.
\newblock Ph.D. thesis, Carnegie Mellon University.

\end{thebibliography}
\bibliographystyle{aaai21}

\appendix

\section{Evaluation Metrics}

For evaluating TP identification, we use three different metrics,
Total Agreement (TA), Partial Agreement (PA) and Distance $D$, as
suggested in \citet{papalampidi2019movie}. TA is the percentage of
TP scenes that are correctly identified, i.e.:
\begin{align}
    TA = \frac{1}{T \cdot L}\sum_{i=1}^{T \cdot L}{\frac{|S_i \cap
      G_i|}{|S_i \cup G_i|}} \label{eq:total_agreement}
\end{align}
where $S_i$ is the set of scenes selected as representative for a specific TP event in a movie, $G_i$ is the ground-truth set of scenes representing this TP event, $T$ is the number of TP events (i.e.,~5) and $L$ is the number of movies contained in the test set. 

PA is the percentage of TP events for which at least one ground-truth TP scene is identified, i.e.:
\begin{align}
    PA = \frac{1}{T \cdot L}\sum_{i=1}^{T \cdot L}{\left[ S_i \cap G_i \neq
      \emptyset \right]} \label{eq:partial_agreement}
\end{align}
Finally, $D$ is the average distance between all pairs of predicted TP scenes $S_i$ and ground-truth TP scenes $G_i$ with respect to the screenplay length (i.e.,~number of screenplay scenes). Formally:
\begin{gather}
    d[S_i, G_i] = \frac{1}{N} \min_{(s \in S_i, g \in G_i)} |s - g|
       \label{eq:distance_script_1} \\
    D = \frac{1}{T \cdot L}\sum_{i=1}^{T \cdot L}{d[S_i,
      G_i]} \label{eq:distance_script_2}
\end{gather}
where $d[S_i, G_i]$ is the minimum distance between two sets $S_i$ and $G_i$ corresponding to the same TP event in a movie.

Intuitively, the TA and PA metrics measure the percentage of exact
matches on TP identification. However, since we seek to identify a
very small number of scenes (i.e., 2--3 on average) in the screenplay
as representative of a specific event, the number of exact matches is
expected to be low. For this reason, we also evaluate model
performance based on the $D$ metric. $D$ gives us an estimate on how
well distributed the identified TP events are in the movie. Moreover,
a large average $D$ indicates that a model is not able to  predict
the correct section in the movie where TP event is located and
hence, such a performance is considered poor.

\begin{table}[t]
\centering
\small
\begin{tabular}{lL{14em}}
\hline 
 \begin{tabular}[c]{@{}c@{}} \textbf{Category} \end{tabular} & \begin{tabular}[c]{@{}c@{}} \textbf{Movies} \end{tabular}  \\ \hline
 \begin{tabular}[c]{@{}l@{}} Comedy/Romance  \end{tabular} & 'Juno', 'The Back-up Plan', 'The Breakfast Club', '500 Days of Summer', 'Crazy, Stupid, Love', 'Easy A', 'Marley \& Me', 'No Strings Attached' \\ \hline
\begin{tabular}[c]{@{}l@{}} Thriller/Mystery  \end{tabular} & 'Arbitrage', 'Panic Room', 'The Shining', 'One Eight Seven', 'Black Swan', 'Gothika', 'Heat', 'House of 1000 Corpses', 'Sleepy Hollow', 'The Talented Mr. Ripley', 'The Thing' \\ \hline
\begin{tabular}[c]{@{}l@{}} Action  \end{tabular} & 'Die Hard', 'Soldier', 'The Crying Game', 'Total Recall',  '2012', 'From Russia with Love', 'American Gangster', 'Collateral Damage','Oblivion' \\ \hline
\begin{tabular}[c]{@{}l@{}} Drama/Other  \end{tabular} & 'Moon', 'Slumdog Millionaire',  'The Last Temptation of Christ', 'Unforgiven', 'American Beauty', 'Jane Eyre',  'The Majestic', 'A Walk to Remember' \\ 
\hline
\end{tabular}
\caption{Movies from test set divided in four broad categories based on their genre.}
\label{tab:movies_per_cat}
\end{table}

\begin{table*}[t]
\centering
\small
\begin{tabular}{cL{27em}L{14em}}
\hline 
 \begin{tabular}[c]{@{}c@{}} \textbf{Movie}  \end{tabular} & \begin{tabular}[c]{@{}c@{}} \textbf{Summary} \end{tabular} & \begin{tabular}[c]{@{}c@{}} \textbf{Questions} \end{tabular} \\ \hline
 \begin{tabular}[c]{@{}l@{}} Arbitrage \end{tabular} & \begin{tabular}[c]{L{27em}} Sixty-year-old magnate Robert Miller manages a hedge fund with his daughter Brooke and is about to sell it for a handsome profit. However, unbeknownst to his daughter and most of his other employees, he has cooked his company's books in order to cover an investment loss and avoid being arrested for fraud. \textcolor{tp1}{One night, while driving with his mistress Julie Cote, he begins to doze off and crashes; Julie is killed.} Miller covers up Julie's death with Jimmy's help. The next day, he is questioned by police detective Bryer. \textcolor{tp2}{Bryer is keen on arresting Miller for manslaughter and begins to put the pieces together.} \textcolor{tp3}{Jimmy is arrested and placed before a grand jury but still refuses to admit to helping Miller.} The case against Jimmy is dismissed and the detective is ordered not to go near him. \textcolor{tp4}{Miller's wife tries to blackmail him with a separation agreement getting rid of his wealth.} \textcolor{tp5}{In the final scene, Miller addresses a banquet honoring him for his successful business, with his wife at his side and his daughter introducing him to the audience but their false embrace on the stage signifies that he has lost the respect and admiration of his daughter.} \end{tabular} &  \begin{tabular}[c]{L{14em}}  1. Did the video summary show the accident and Julie's death? \\ \\ 2. Did the video summary show Bryer's suspicions towards Miller? \\ \\ 3. Did the video summary show Jimmy's arrest and the police's efforts to get Jimmy to admit helping Miller? \\ \\ 4. Did the video summary show that Miller's wife blackmailed him? \\ \\ 5. Did the video summary show Miller give a speech about his successful business with his family on his side in the end? \end{tabular} \\ \hline
 
  \begin{tabular}[c]{@{}l@{}} The Back-up Plan \end{tabular} & \begin{tabular}[c]{L{27em}} Zoe has given up on finding the man of her dreams and decided to become a single mother and undergoes artificial insemination. \textcolor{tp1}{The same day she meets Stan when they both try to hail the same taxi.} After running into each other twice more, Stan asks Zoe on a date. \textcolor{tp2}{At the end of the date Stan asks her to come to his farm during the weekend and Zoe finds out that she is pregnant.} In the farm Zoe tells Stan that she is pregnant and he is confused and angry at first. \textcolor{tp3}{However, Stan decides he still wants to be with her and they reconcile.} After many misunderstandings and comedic revelations, Zoe and Stan are walking into the Market when they run into Stan's ex-girlfriend. \textcolor{tp4}{Due to Stan's remark that the twins are not his, Zoe believes that he is not ready to become a father to them, and breaks off the relationship.} At her grandmother's wedding, Zoe's water breaks and on the way to the hospital they make a pit stop at the Market. Zoe apologizes to Stan and they begin to work things out. \textcolor{tp5}{He pulls out a penny, that he kept from their first acquaintance, and Zoe promises to trust him more.} \end{tabular} &  \begin{tabular}[c]{L{14em}}  1. Did the video summary show Zoe and Stan's first encounter in a taxi? \\ \\ 2. Did the video summary show Zoe discover that she is pregnant? \\ \\ 3. Did the video summary show Stan's commitment to Zoe after finding out that she is pregnant? \\ \\ 4. Did the video summary show Zoe and Stan's break up? \\ \\ 5. Did the video summary show the final reconciliation between Stan and Zoe? \end{tabular} \\ \hline
 \begin{tabular}[c]{@{}l@{}} Juno \end{tabular} & \begin{tabular}[c]{L{27em}} \textcolor{tp1}{Sixteen-year-old Minnesota high-schooler Juno MacGuff discovers she is pregnant with a child fathered by her friend and longtime admirer, Paulie Bleeker.} She initially considers an abortion, but \textcolor{tp2}{finally decides to give the baby up for adoption.} \textcolor{tp3}{Juno meets a couple, Mark and Vanessa Loring, in their expensive home and agrees to a closed adoption.} After some time, where Juno gets to know the couple better and not long before her baby is due, she is again visiting Mark when their interaction becomes emotional. A few moments later, \textcolor{tp4}{she watches the Loring marriage fall apart, then drives away and breaks down in tears by the side of the road.} Not long after, Juno goes into labor and is rushed to the hospital, where she gives birth to a baby boy. \textcolor{tp5}{Vanessa comes to the hospital where she joyfully claims the newborn boy as a single adoptive mother.} \end{tabular} &  \begin{tabular}[c]{L{14em}}  1. Did the video summary show Juno discover that she is pregnant? \\ \\ 2. Did the video summary show Juno's decision to give the baby up for adoption? \\ \\ 3. Did the video summary show Juno's first visit to Mark and Vanessa's house? \\ \\ 4. Did the video summary show Mark and Vanessa's breakup? \\ \\ 5. Did the video summary show Vanessa adopt Juno's baby as a single mother in the end? \end{tabular} \\ \hline
\end{tabular}
\caption{Examples of movies used for human evaluation. We present a short text summary per movie based on Wikipedia and the questions we asked for evaluating the information contained in each video summary. Different colors in the summary correspond to different TPs (i.e., \textcolor{tp1}{TP1}, \textcolor{tp2}{TP2}, \textcolor{tp3}{TP3}, \textcolor{tp4}{TP4}, \textcolor{tp5}{TP5}).}
\label{tab:amt_examples_1}
\end{table*}

\begin{figure*}[h!]
    \centering
\begin{tabular}{|c|c|c|} \hline
        {{\small The Wedding Date}} & {{\small Easy A}} & {{\small Marley \& Me}}  \\  
         \includegraphics[width=0.3\textwidth]{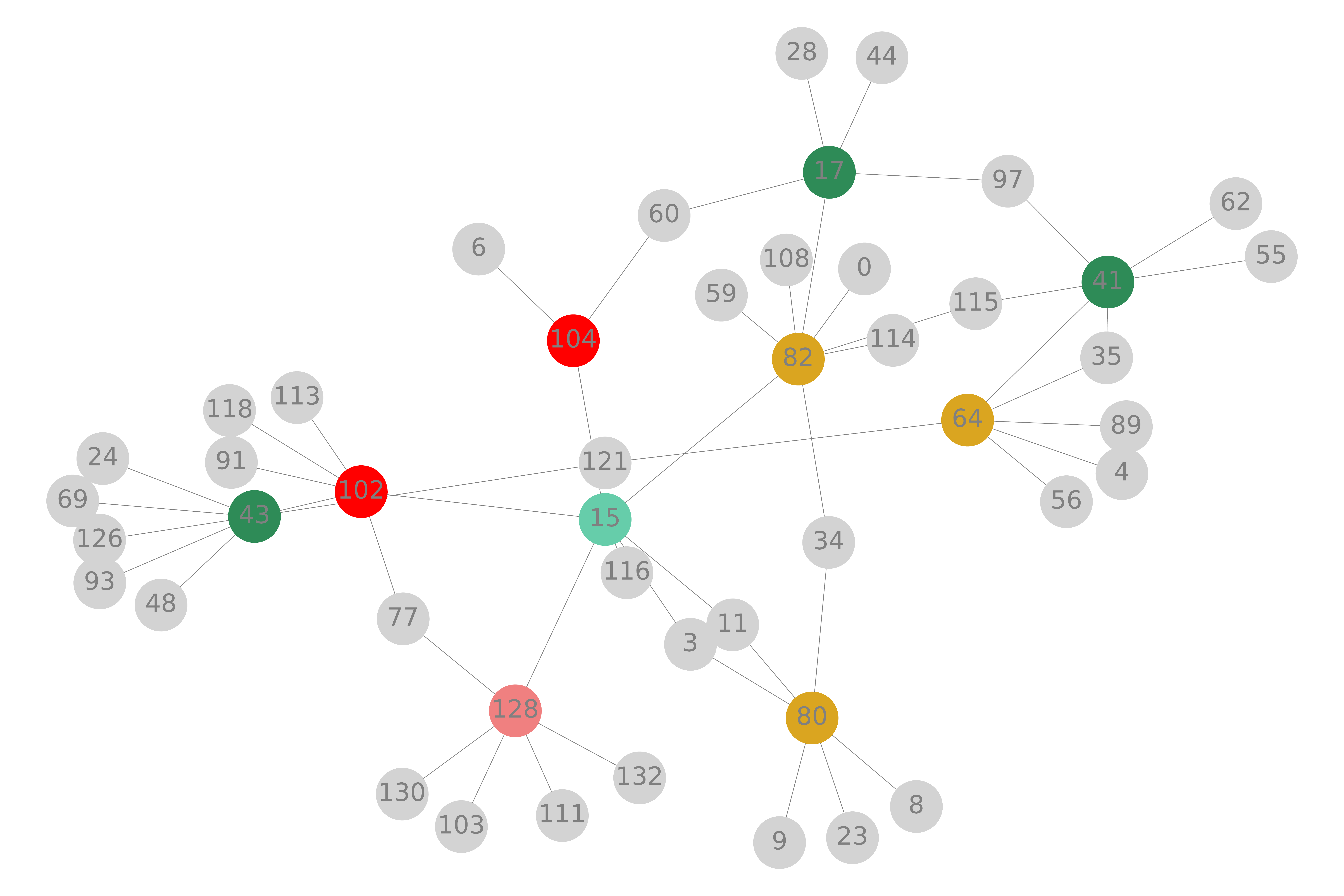} &          
         \includegraphics[width=0.3\textwidth]{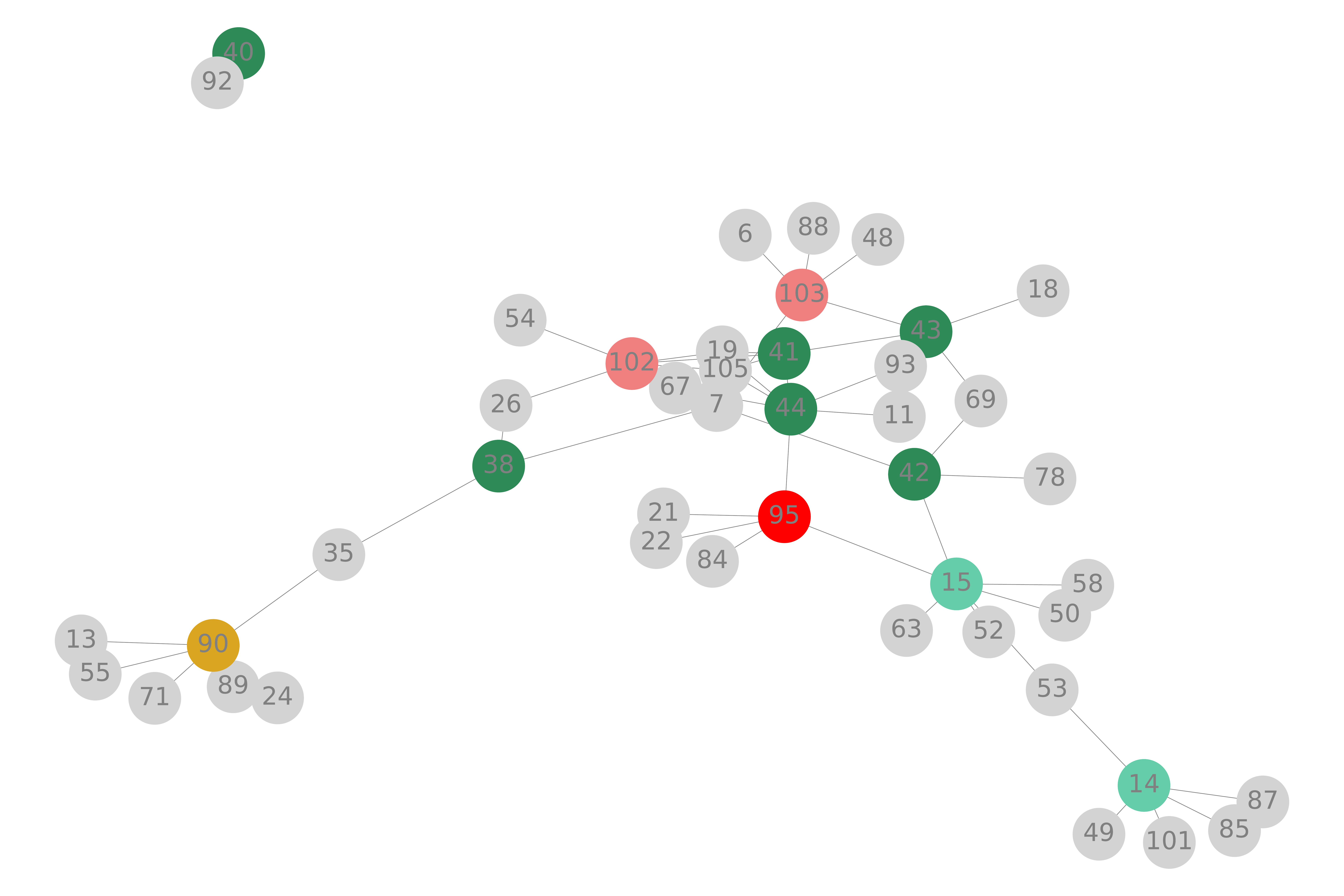} &
        \includegraphics[width=0.3\textwidth]{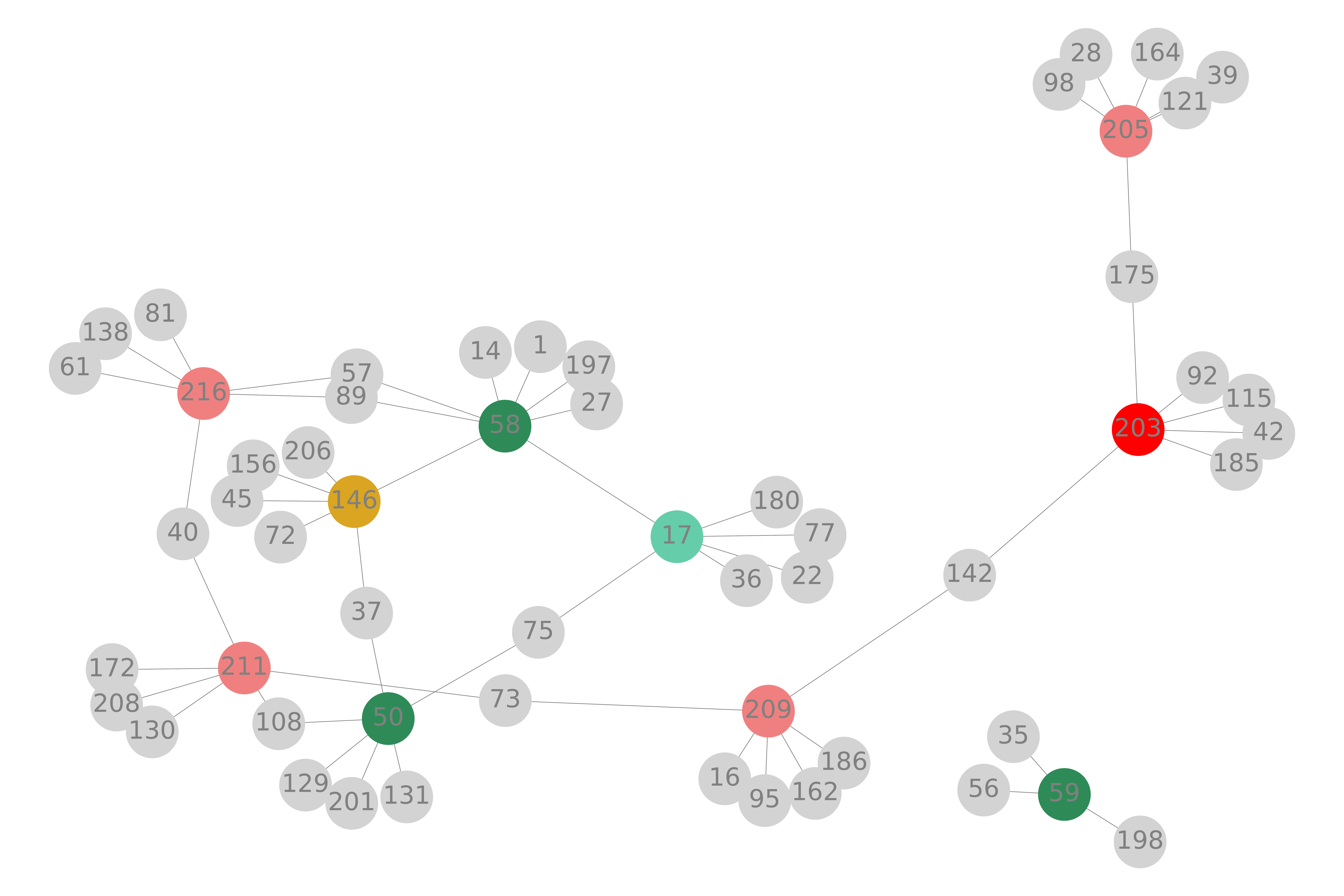} \\ \hline
        
        {{\small Die Hard}} & {{\small Total Recall}} & {{\small From Russia With Love}}  \\ 
        \includegraphics[width=0.3\textwidth]{graph_analysis/die_hard_graph_w_labels.pdf}& 
        \includegraphics[width=0.3\textwidth]{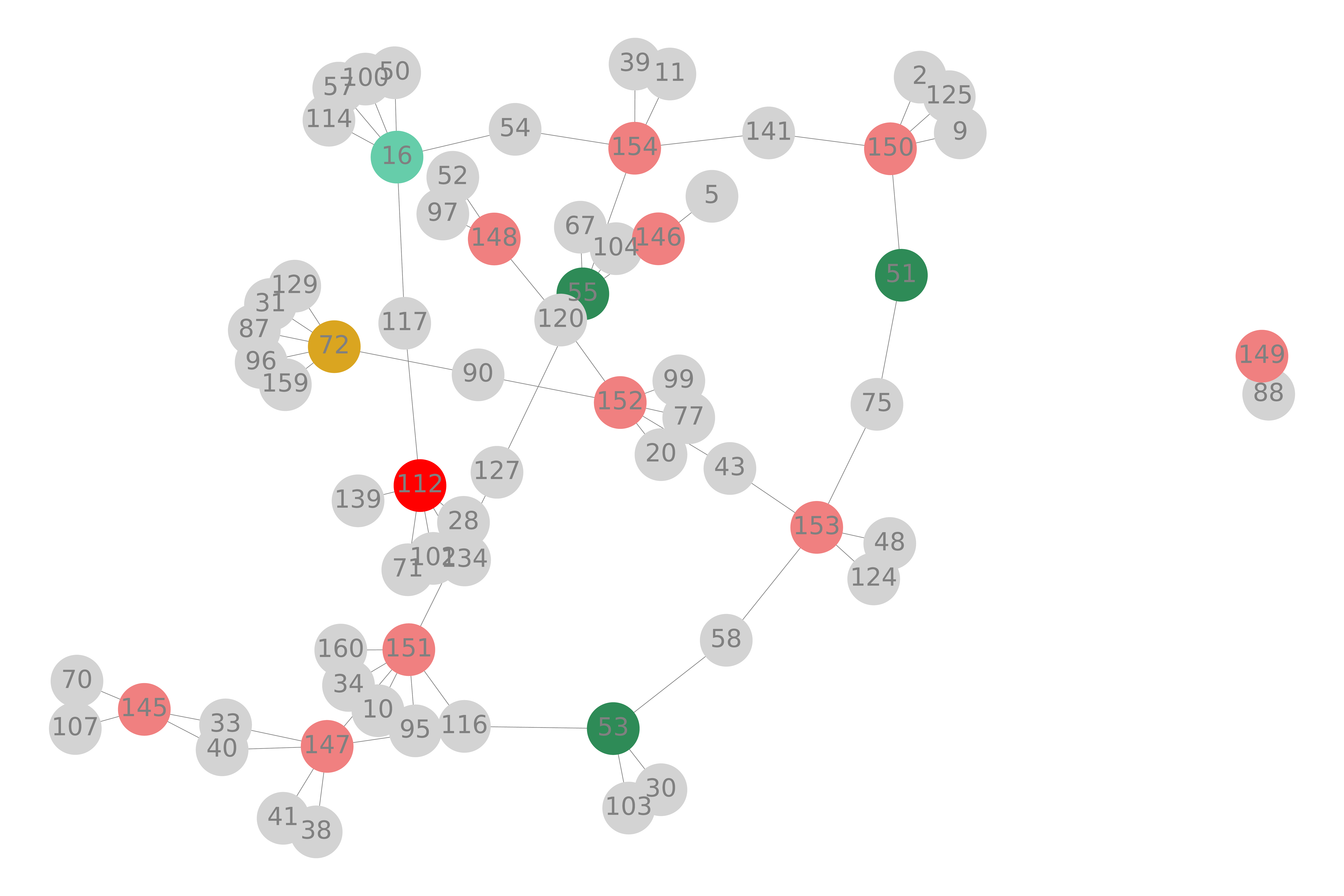} &
        \includegraphics[width=0.3\textwidth]{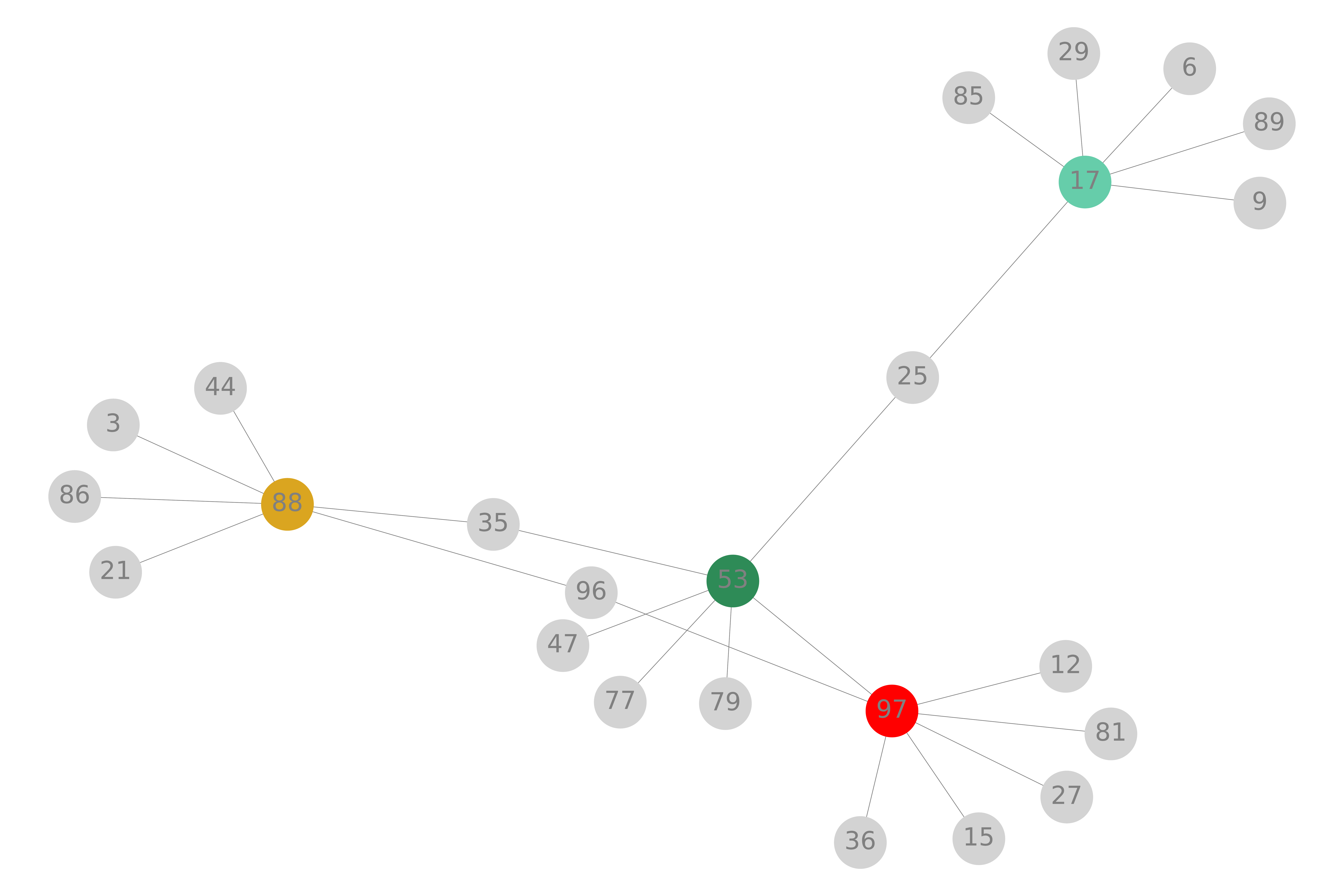}\\\hline
        
\end{tabular}
 \caption{Examples of graphs produced by \graphtp (+audiovisual) for movies from the test set. Nodes  (in color) are scenes which
   act as TPs and their immediate neighbors (in gray). Genre category per row: comedy/romance, action. All graphs are dense and present high connectivity.
   \label{fig:graph_examples_complete_1}}
\end{figure*}

\begin{figure*}[t]
    \centering
\begin{tabular}{|c|c|c|} \hline
        {{\small Panic Room}} & {{\small The Thing}} & {{\small Black Swan}}  \\  
         \includegraphics[width=0.3\textwidth]{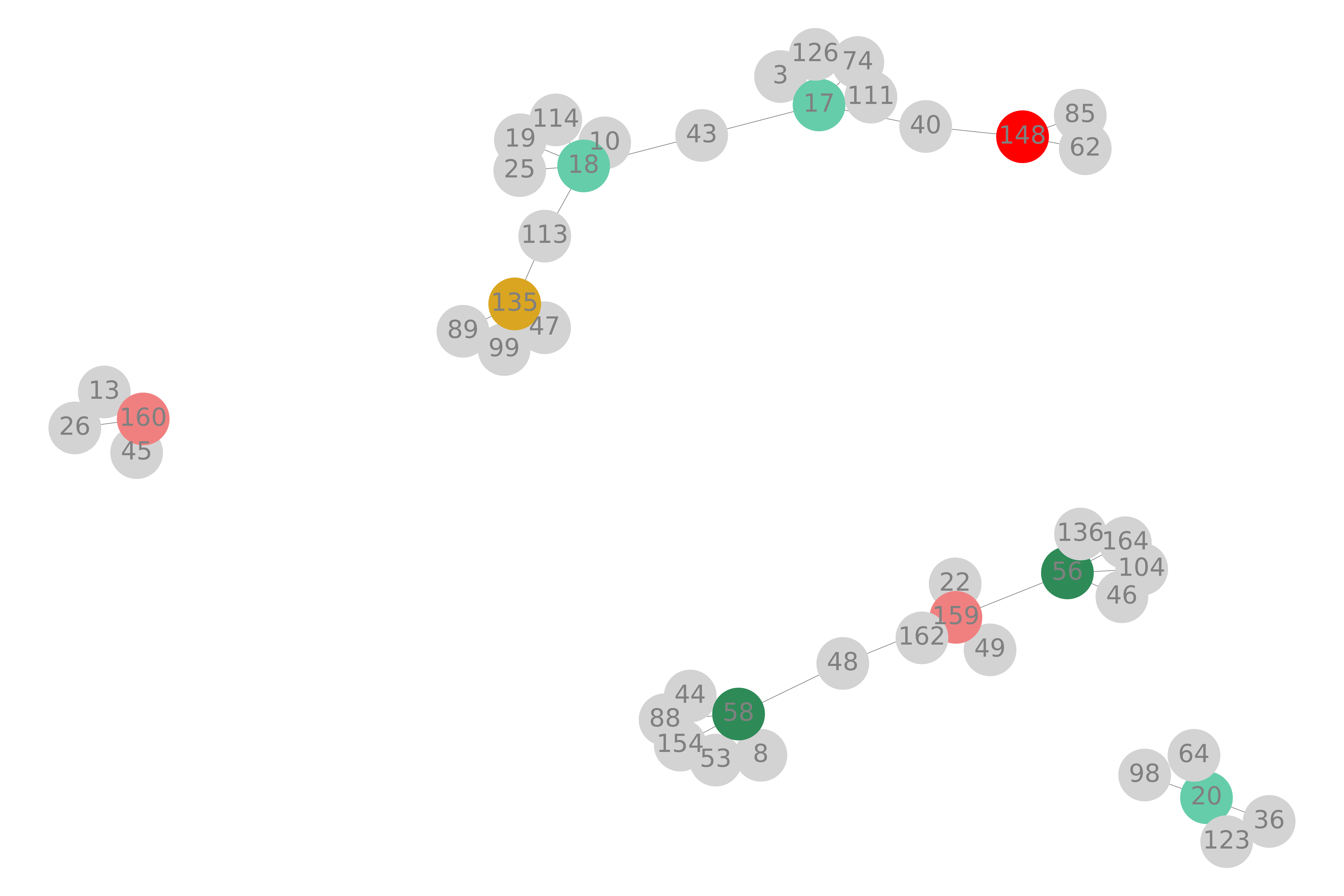} &          
         \includegraphics[width=0.3\textwidth]{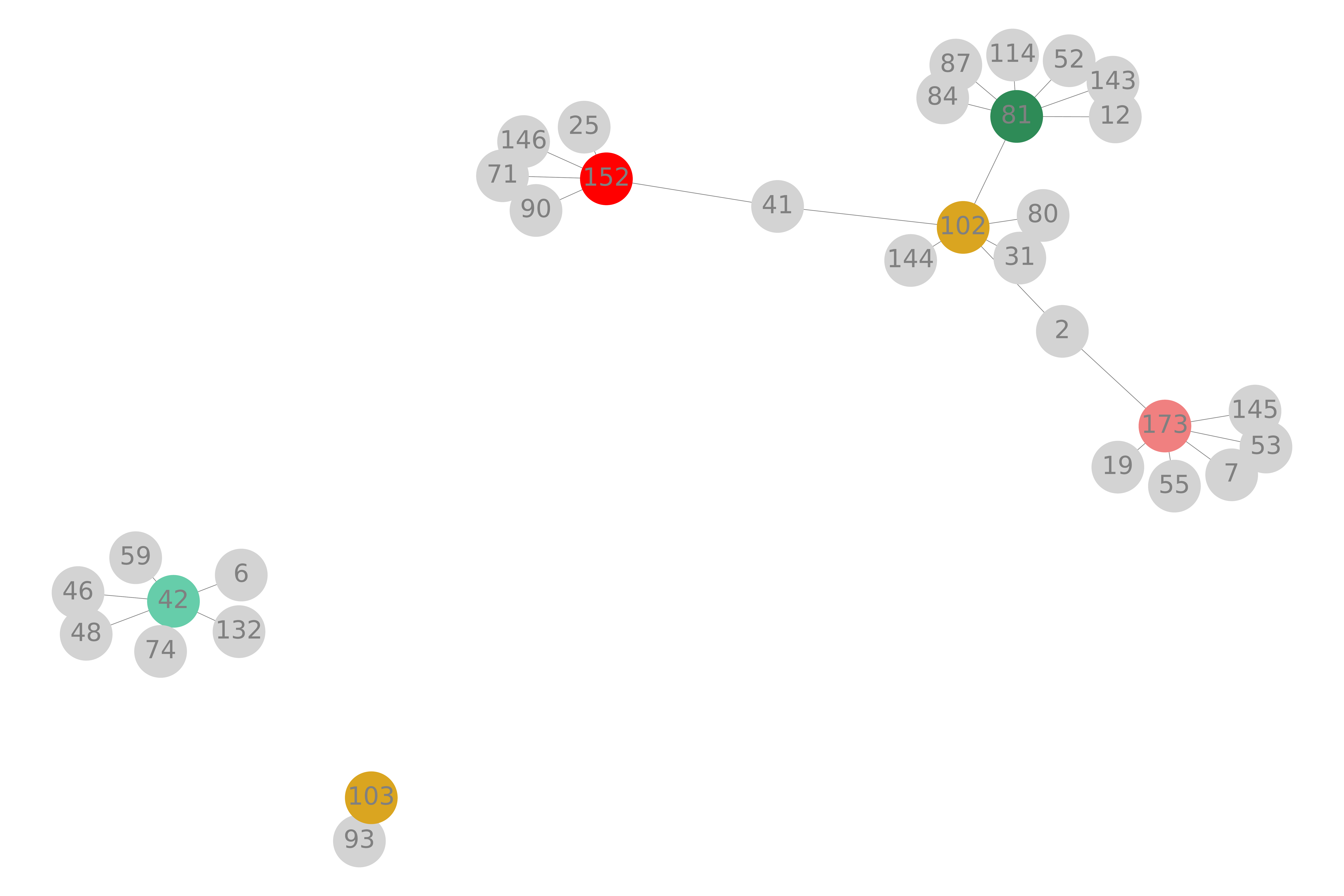} &
        \includegraphics[width=0.3\textwidth]{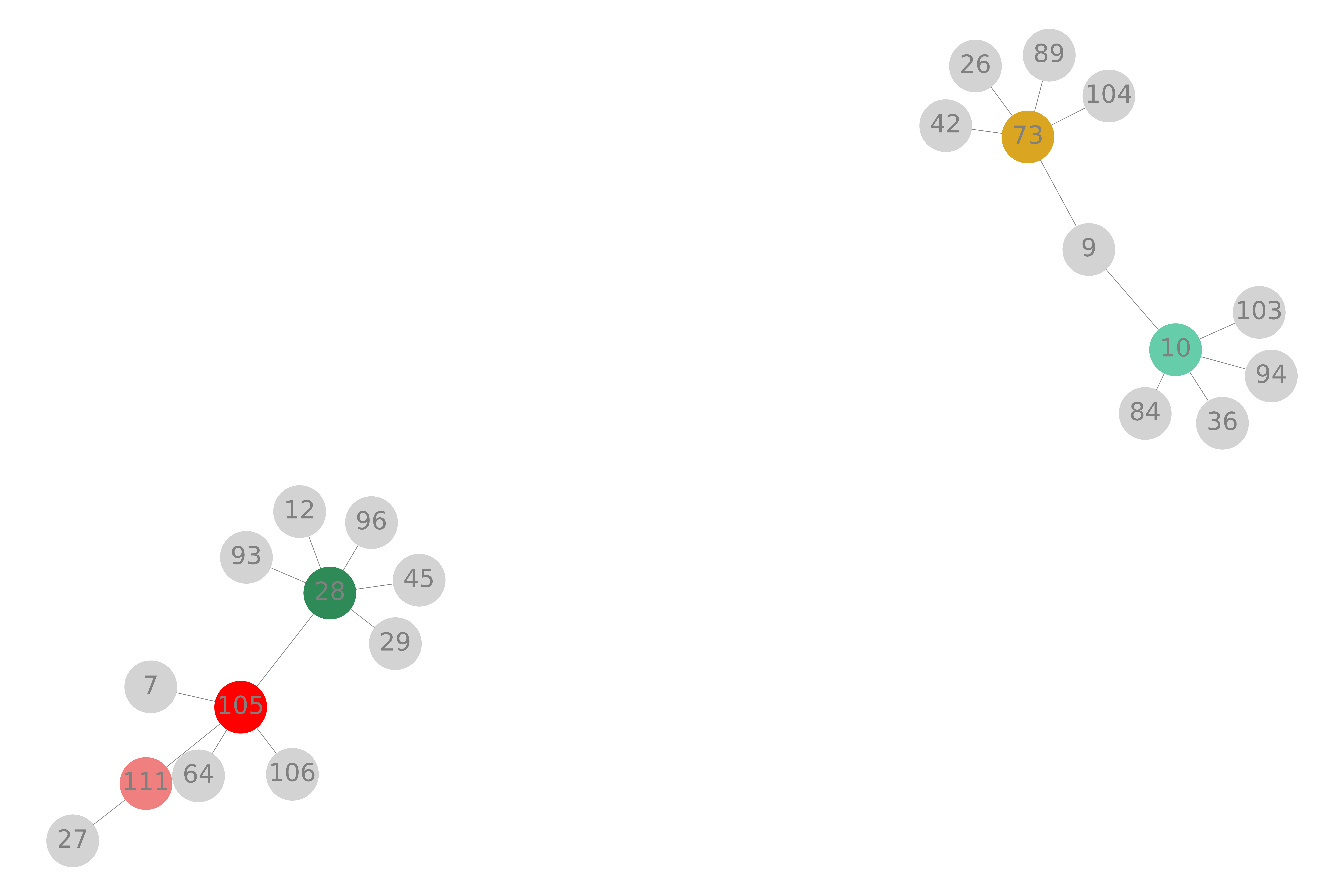} \\ \hline
        
        {{\small American Beauty}} & {{\small Jane Eyre}} & {{\small Slumdog Millionaire}}  \\ 
        \includegraphics[width=0.3\textwidth]{graph_analysis/american_beauty_graph_w_labels.pdf}& 
        \includegraphics[width=0.3\textwidth]{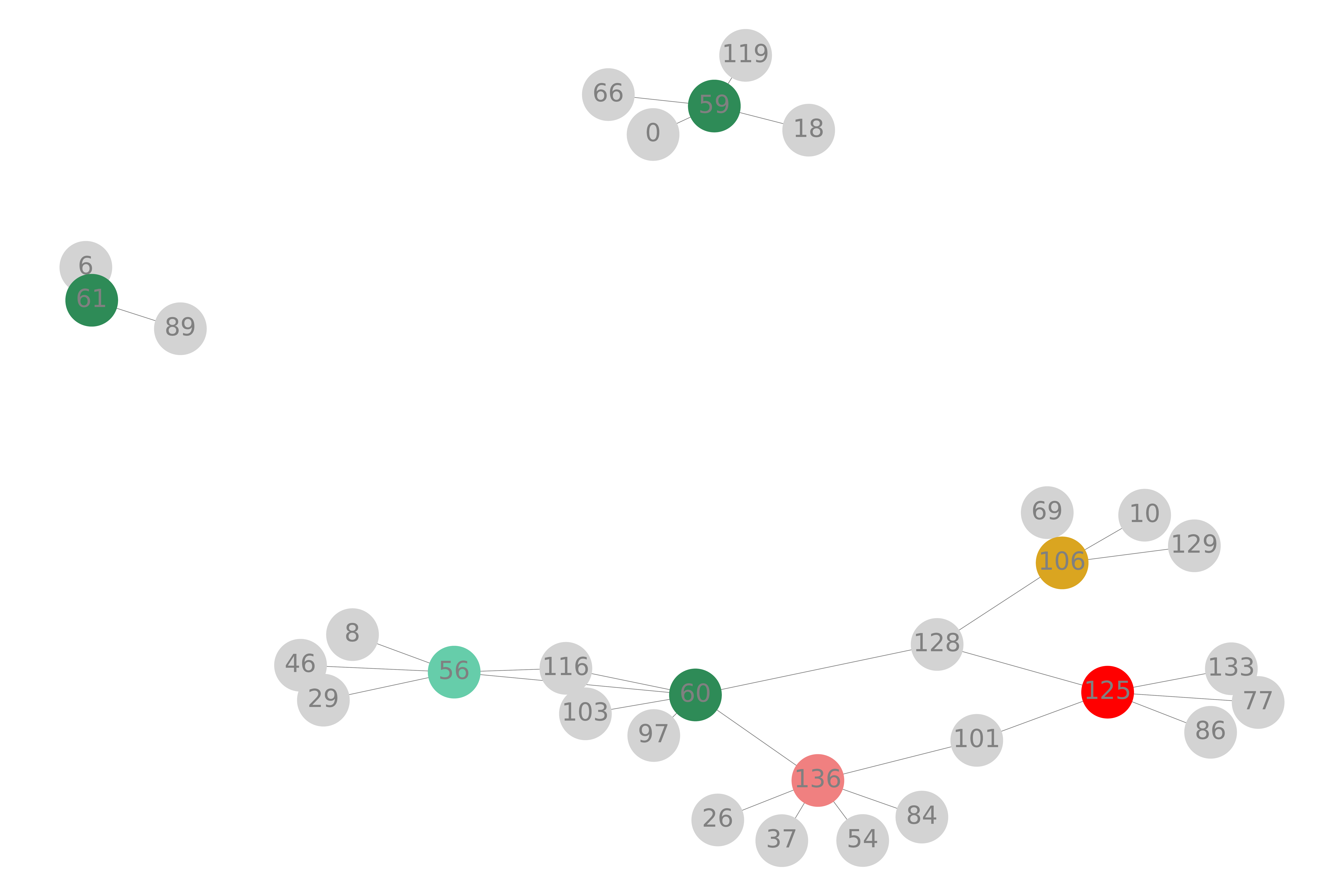} &
        \includegraphics[width=0.3\textwidth]{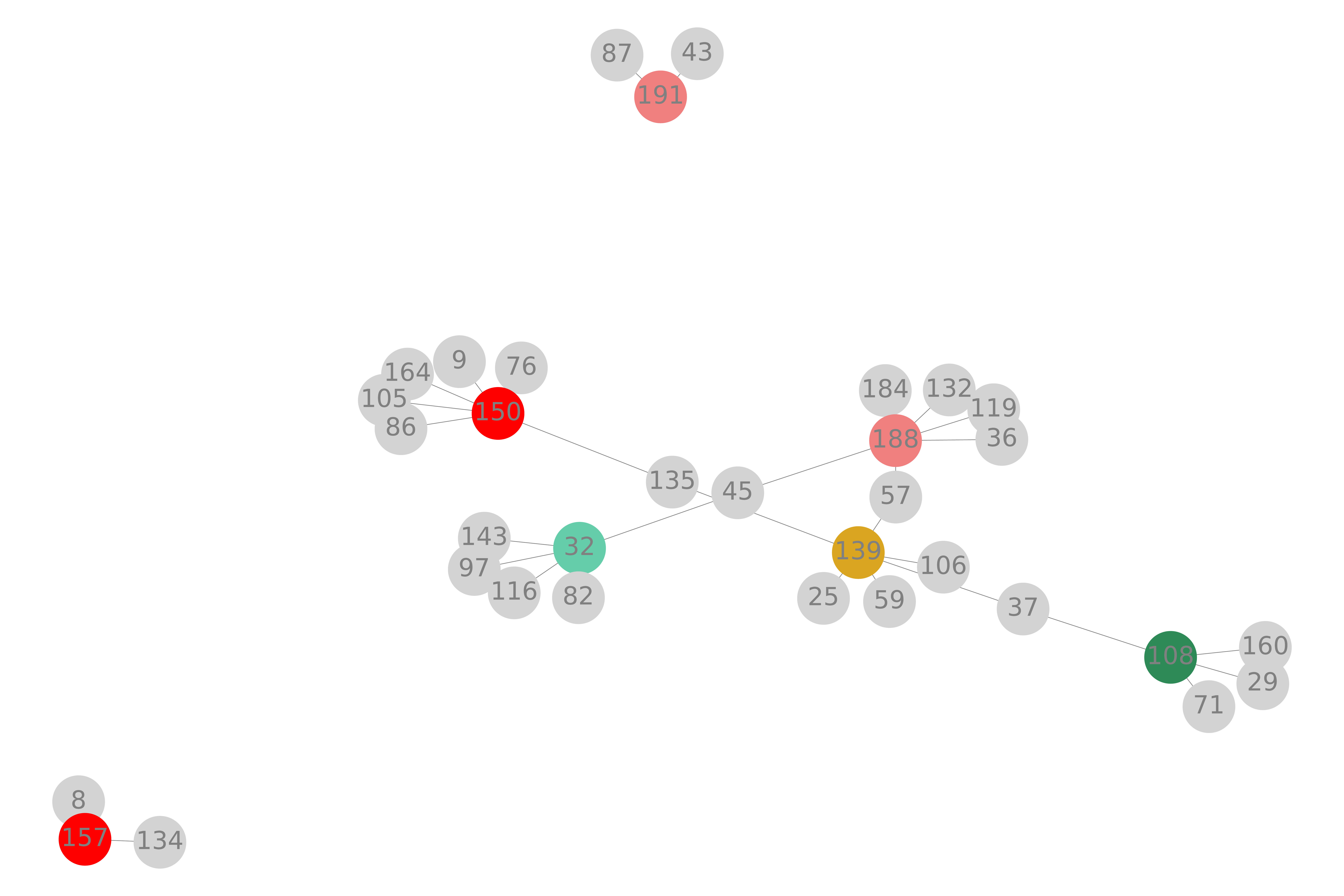}\\\hline
\end{tabular}
 \caption{Examples of graphs produced by \graphtp (+audiovisual) for movies from the test set. Nodes  (in color) are scenes which
   act as TPs and their immediate neighbors (in gray). Genre category per row: thriller/mystery, drama/other. The graphs here present low connectivity containing several disconnected subgraphs. 
   \label{fig:graph_examples_complete_2}}
\end{figure*}

\begin{figure*}[t]
    \centering
    \includegraphics[width=\textwidth]{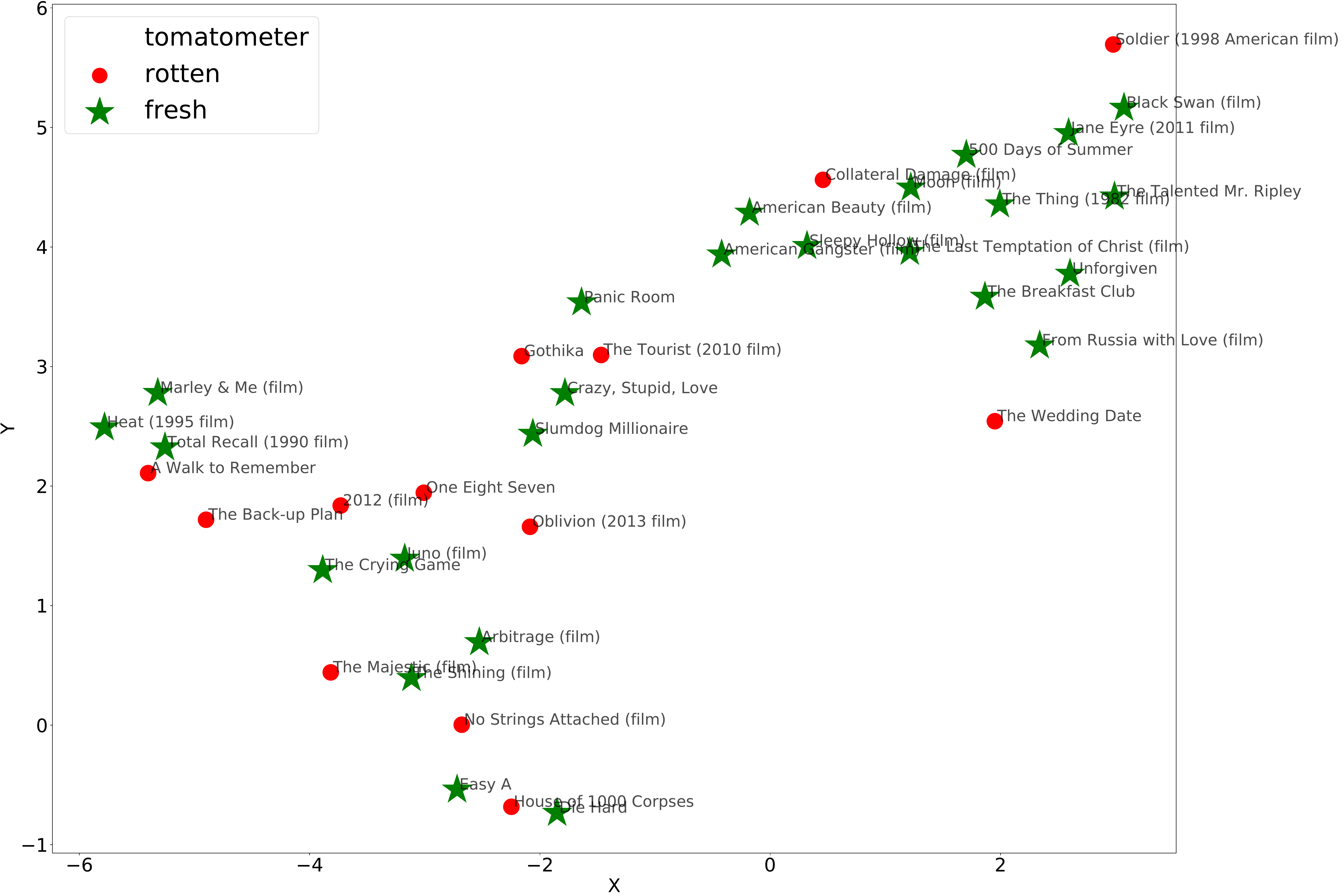}
        \caption{Analysis of the graphs for the movies of the test set with regard to their scores in Rotten Tomatoes. We computed various metrics, such as node connectivity, diameter and triadic closure, that describe each movie graph. Next, we performed dimensionality reduction using UMAP and visualize all movies in two dimensions. A small distance between movies in the figure indicates structural similarity between their graphs. Regarding the scores, rotten are the movies with a score lower than 60\%, otherwise they are fresh.
    }
    \label{fig:tomatometer}%
\end{figure*}

\section{Comparison Systems}

Except for \graphtp, we evaluate the performance of the general summarization algorithm \textrank, its character-based version \scenesum specific to narratives and a sequence-based TP identification model, namely TAM.

\subsection{\textrank}

For \textrank, we consider a screenplay graph $\mathcal{G}=(\mathcal{V}, \mathcal{E})$, with nodes $\mathcal{V}$ representing the scenes and edges $\mathcal{E}$ the similarity between scenes, similarly with \graphtp. We compute the textual scene representations $s$ and the similarity $e_{ij}$ between all pairs of scenes $s_i, s_j$ as described in \citet{papalampidi2020screenplay}. Finally, for each scene $s_i$ we compute a centrality score:
\begin{gather}
    \textit{centrality}(s_i) = \lambda_1  \sum_{j<i}e_{ij} + \lambda_2  \sum_{j>i}e_{ij} \label{eq:centrality_textrank}
\end{gather}
where $\lambda_1=\lambda_2=0.5$. We use the undirected version of
\textrank, since when considering different values for $\lambda_1$ and
$\lambda_2$ \cite{zheng2019sentence,papalampidi2020screenplay}, only
an isolated part of the movie receives high centrality
scores. Finally, we select the five scenes with the highest centrality
scores and sequentially assign them to each TP. In accordance with
\graphtp, each TP is again represented by three consecutive scenes
(the currently selected scene and its immediate neighbors in the
screenplay).

For the multimodal version of \textrank we again compute the
audiovisual similarity $u_{ij}$ between all pairs of scenes
$s_i, s_j$, as for \graphtp. The only difference here is that we do
not use trainable layers to project the audio and visual
representations for the segments and frames, respectively and we do
not consider attention weights when combining different
representations, since \textrank is an unsupervised algorithm with no
training signal. Finally, we again combine textual~$e_{ij}$ and
audiovisual $u_{ij}$ similarities via multiplication. The combined
similarity score is now used in Eq.~\eqref{eq:centrality_textrank}.

\subsection{\scenesum}

\scenesum shares the same architecture with \textrank. However, it also considers the characters participating in each scene as initially suggested by \citet{gorinski-lapata-2015-movie} and later adopted by \citet{papalampidi2020screenplay}.

Our implementation of \scenesum follows \citet{papalampidi2020screenplay}. For each scene $s_i$ we calculate a character-related importance score $c_i$:
\begin{gather}
    c_i = \frac{\sum_{c \in C}\ [c \in C_i\ \cup\
      \textit{main}(C)]}{\sum_{c \in C}\ [c \in C_i]}
 \label{eq:char_scores}
\end{gather}
where $C_i$ is the set of all characters participating in scene $s_i$,
$C$ is the set of all characters participating in the movie and
$\textit{main}(C)$ are all the main characters. \citet{gorinski-lapata-2015-movie} retrieve the main characters in a movie by constructing a social network depicting all interactions between characters and select the most central ones as the protagonists. We follow a simpler approach and consider as main characters the ones that are mentioned in the respective Wikipedia synopsis. We add the importance scores $c_i$ and $c_j$ for scenes $s_i$ and $s_j$ to the similarity score $e_{ij}$ --or the combined multimodal score $u_{ij}*e_{ij}$-- in Eq. \eqref{eq:centrality_textrank} via summation \cite{papalampidi2020screenplay}. 

\subsection{Topic-Aware Model (TAM)}

TAM is originally proposed by \citet{papalampidi2019movie} for
identifying scenes in a screenplay that describe a synopsis sentence
marked as TP. Later, \citet{papalampidi2020screenplay} modified TAM
to only consider screenplay scenes --no synopsis information-- for
predicting TP events. In this work, we use the later modification of
TAM with the same implementation details. TAM operates over a sequence
of contextualized scene representations derived from a BiLSTM over the
whole screenplay. In order to better capture interactions between
scenes TAM considers a sliding context window of fixed length and
computes the similarity of a current scene with a previous and next
context representation. The hypotheses for TAM are that (1) TPs act as
boundaries between sections and hence a plunge in the similarity
computation should be noticed close to a TP and (2) the screenplay
presents a sequence of events; no entanglement between stories and
events is taken into account.  For the multimodal version of TAM, we
compute audio and visual representations for each scene as for
\graphtp. Next, we combine the different modalities for the scene via
concatenation.

\section{Implementation Details}

\graphtp has 467.1K trainable parameters when using only textual information and 797.2K when adding the audiovisual features. Similarly, text-only TAM has 401.2K and multi-modal TAM has 731.7K trainable parameters.
We trained models on a single GPU (GeForce GTX 1080). The average runtime is 20 minutes for training the text-only models and 3 hours for the multi-modal ones.

\section{Human Evaluation}

For the human evaluation experiment in Amazon Mechanical Turk (AMT),
we created textual summaries describing a movie's storyline from
beginning to end. Specifically, we produced a shorter version of the
Wikipedia plot synopsis for each movie keeping only essential
information. Next, we colored differently the text in the summary that
corresponded to each TP in order to clearly demonstrate the important
events in the movie.

We asked AMT workers to first read this textual summary paying extra attention to the descriptions in colored text. Next, they watched an automatically produced video summary for the movie and were asked to answer to five questions. Each question examined whether a specific TP event was presented in the video summary. Examples of the summaries given to the AMT workers alongside with the corresponding questions are presented in Table \ref{tab:amt_examples_1}. Finally, we asked them to provide us with an overall rating for the video summary. In this point, we asked workers to take into account the questions answered previously, but also consider the quality of the summary (i.e., how condensed the summary was, whether it contained redundant events, overall information provided). The rating scale was 1 to 5, with 5 being the most informative and least redundant. 

\section{Graph Analysis}

Our evaluation set contains 38 movies annotated with gold-standard scene-level TP labels. During the analysis of the movie graphs, we divide the movies of the test set into four broad categories based on their genre: comedy/romance, thriller/mystery, action and drama/other. In Table \ref{tab:movies_per_cat}, we present how these movies are categories into genres. When a movie is characterised by more than one genres, we categorize it only in its main category.

As described in Section 6, we create pruned graphs for each movie that contain only the nodes that act as TPs and their immediate neighbors. We consider these pruned graphs as a graphical description of a movie's storyline containing only the most important events and their semantic connections. In Figures~\ref{fig:graph_examples_complete_1} and~\ref{fig:graph_examples_complete_2} we present examples of such graphs for various movies of the test set. Figure~\ref{fig:graph_examples_complete_1} contains movies belonging to the comedy/romance and action genre categories per row, respectively. Similarly, Figure~\ref{fig:graph_examples_complete_2} presents illustrations of graphs for thriller/mystery and drama/other movies per row. We can empirically observe that comedies and action movies (Figure~\ref{fig:graph_examples_complete_1}) tend to present denser graphs in comparison with thrillers and dramas (Figure~\ref{fig:graph_examples_complete_2}) which contain several disconnected subgraphs. 

Except for each movie's genre, we also consider the quality of the movie when analyzing its graph. Specifically, we hypothesize that movies with high ratings (e.g. according to Rotten Tomatoes\footnote{\url{https://www.rottentomatoes.com/}}) often have less predictable and distinctive storylines and hence would present a different graph topology in comparison with those of poorly rated movies. In order to test this hypothesis, we first analyze the graph topology using several metrics from graph theory. Specifically, as when examining the graphs with regard to different genres, we again consider the average node connectivity and the node connectivity per TP.  However, we now also compute four extra metrics: the TP pairwise node connectivity (i.e.,~the average node connectivity when considering only the nodes that act as TPs in order to directly measure the degree of connectivity between TP events), the number of disconnected components presented in the graph, its diameter (i.e.,~the greatest distance between any pair of nodes) and the triadic closure (i.e.,~tendency of edges to form triangles). Next, we perform dimensionality reduction using UMAP \cite{mcinnes2018umap} in order to project the movies of the test set to two dimensions and visualize them. 

We present the visualization of the movies in two dimensions in Figure \ref{fig:tomatometer}. Movies are colored differently depending on the score that they received in Rotten Tomatoes (i.e.~movies with a score equal or higher than 60\% are considered fresh (green star), otherwise rotten (red circle)). Although movies are not perfectly divided into clusters according to their score category, we observe that there is a tendency for movies with high scores to concentrate in the upper right side of the figure, whereas a lot of poorly rated movies are located in the leftmost, down side. This suggests that although each movie has a unique storyline, and hence graph, the way that important events are connected plays a significant role to the quality of the movie.  Notably, a lot of well-known unconventional movies with out-of-the-box storylines are located in the ''fresh'' cluster in the figure: e.g.,~''American Beauty'', ''Black Swan'', ''The Thing', ''Moon''. 

\end{document}